\documentclass[onecolumn,10pt]{article}
\usepackage{cite,footnote,xspace,syntonly}
\usepackage{amsmath,amssymb,amsfonts,amsthm,makeidx,graphicx,colortbl}
\newcommand{\pinv}{^{\dagger}}

\newtheorem{mydefinitionhere}{Definition}

\newcommand{\bdiag}[1]{\mathop{\rm bdiag}\brackets{#1}}

\newcommand{\subjectto}{\mathop{\text{s.t.}}}
\newcommand{\columnspan}{\hc{\mathcal{R}}}
\newcommand{\samplerkhsvec}{\hct{\bar{\bm \rkhsfunsymbol}}}
\newcommand{\hct}[1]{#1} 
\newcommand{\diag}[1]{\mathop{\rm diag}\brackets{#1}}

\newcommand{\noisesamp}{{\hc{e}}} 
\newcommand{\noisevec}{\hc{\bm \noisesamp}} 
\newcommand{\noisevar}{\hc{ {\sigma^2_\noisesamp}}}  %
\newcommand{\samplingmat}{\hc{\boldmatvec S}}

\newcommand{\rfield}{\mathbb{R}}
\newcommand{\vect}{\mathop{\rm vec}}
\newcommand{\kernelmap}{\hc{\kappa}}

\newcommand{\hbm}[1]{{\hat{\bm #1}}}
\newcommand{\inv}{^{-1}}

\newcommand{\adjtransgraphmat}{\hc{\bm A}}
\newcommand{\adjacencytildmat}{\hc{ \tilde{\bm A}}}

\newcommand{\plantnoisetildvec}{\hc{\tilde{\bm{\plantnoise}}}}
\newcommand{\truesignalspatiotempcompfun}{\hc{ f}^{(\spatiotempind)}}
\newcommand{\truesignalspatiocompfun}{\hc{ f}^{(\spatioind)}}
\newcommand{\timeperiodgiventimenot}[1]{\hc{_{{1:\timenum}|{#1}}}}
\newcommand{\expectednb}{\hc{\mathbb{E}}}
\newcommand{\expected}[1]{\expectednb\left[#1\right]}  

\newcommand{\bm}[1]{\hc{\boldsymbol{\mathbf{#1}}}}

\newcommand{\gausmeanspatio}{\hc{\mu}_\spatioind}
\newcommand{\gausmeanstatenoise}{\hc{\mu}_\plantnoise}
\newcommand{\gausstdspatio}{\hc{r}_\spatioind}
\newcommand{\gausstdstatenoise}{\hc{r}_\plantnoise}

\newcommand{\fullalphavec}{\hct{\bm \upalpha}} 
\newcommand{\transweight}{\hc{\alpha}}

\newcommand{\plantnoiseweight}{\hc{s_\plantnoise}}
\newcommand{\rkhsspationum}{{\hc{M}_\spatioind}}
\newcommand{\rkhsstatenoisenum}{{\hc{M}_\plantnoise}}

\newcommand{\kernelcoefnot}[1]{\hc{(} {#1}\hc{)}}
\newcommand{\kernelcoefspatioreg}{{\hc{\rho}_\spatioind}}
\newcommand{\kernelcoefstatenoisereg}{{\hc{\rho}_\plantnoise}}

\newcommand{\kernelcoefspatiovec}{\hc{\boldmatvecgreek \uptheta}^{(\spatioind)}}
\newcommand{\kernelcoefspatio}{\hc{\theta}^{(\spatioind)}}
\newcommand{\kernelcoefspatiovecest}{\hc{\hat{\boldmatvecgreek \uptheta}}^{(\spatioind)}}
\newcommand{\kernelcoefstatenoisevec}{\hc{\boldmatvecgreek \uptheta}^{(\plantnoise)}}
\newcommand{\kernelcoefstatenoise}{\hc{\theta}^{(\plantnoise)}}
\newcommand{\kernelcoefstatenoisevecest}{\hc{\hat{\boldmatvecgreek \uptheta}}^{(\plantnoise)}}
\newcommand{\fullkernelspatiomatdict}{\hc{\mathcal{D}}_\spatioind}

\newcommand{\fullkernelstatenoisematdict}{\hc{\mathcal{D}}_\plantnoise}

\newcommand{\fullkernelspatiomat}{\hc{\fullkernelmat}^{(\spatioind)}}

\newcommand{\fullkernelstatenoisemat}{\hc{\fullkernelmat}^{(\plantnoise)}}

\newcommand{\regparone}{\hc{\mu_1}}
\newcommand{\regpartwo}{\hc{\mu_2}}

\newcommand{\genkernelmat}{\hc{\reducedkernelmat}^{(\spatiotempind)}}
\newcommand{\reducedkernelmat}{\hc{\bar{\boldmatvec K}}}

\newcommand{\kernelindnot}[1]{{{\hc{(}{#1}\hc{)}}}}  %

\newcommand{\define}{:=}
\DeclareMathOperator*{\argmin}{arg\,min}

\newcommand{\boldmatvec}{\mathbf}
\newcommand{\boldmatvecgreek}{\bm}

\newcommand{\transpose}{^{\hc{T}}}
\newcommand{\bandwidth}{\hc{B}}
\newcommand{\truesignal}{\hc{\boldmatvec f}}
\newcommand{\spatiotempind}{\chi}
\newcommand{\spatioind}{\nu}

\newcommand{\truesignalspatiotempcomp}{\hc{\boldmatvec f}^{(\spatiotempind)}}

\newcommand{\truesignalspatiocomp}{\hc{\boldmatvec f}^{(\spatioind)}}

\newcommand{\estsignalspatiotempcomp}{\hc{ \hat{\boldmatvec f}}^{(\spatiotempind)}}
\newcommand{\estsignalspatiocomp}{\hc{ \hat{\boldmatvec f}}^{(\spatioind)}}

\newcommand{\projectmat}{\hc{\boldmatvec P}}

\newcounter{exampleind}

\newcounter{remarkind}

\newcommand{\canonicalvec}[2]{{\bm i}_{#1,#2}}
\newcommand{\tridiag}{\mathop{\rm btridiag}}

\newcommand{\pdset}{\hc{\mathbb{S}}_+}
\newcommand{\identitymat}{\hc{\boldmatvec I}}

\newcommand{\timenot}[1]{_{#1}}
\newcommand{\timescalarnot}[1]{\hc{(}{#1}\hc{)}}
\newcommand{\timegiventimenot}[2]{_{{#1}|{#2}}}  %
\newcommand{\timegiventimevertexnot}[3]{_{#3}{\hc{(}{#1}|{#2}\hc{)}}}  %
\newcommand{\timetimenot}[2]{_{{#1},{#2}}}  %
\newcommand{\timeind}{{\hc{t}}} 
\newcommand{\timeindaux}{{\hc{\tau}}} 
\newcommand{\timeindp}{{\hc{\timeind}'}} 
\newcommand{\timenum}{{\hc{T}}} 
\newcommand{\timeset}{\hc{\mathcal{T}}} 
\newcommand{\timevertexnot}[2]{{\hc{(}v_{#2},{#1}\hc{)}}}  
\newcommand{\vertextimenot}[2]{_{#2}{\hc{(}{#1}\hc{)}}}  %
\newcommand{\vertexvertexnot}[2]{_{#1,#2}}  %
\newcommand{\timevertexvertexnot}[3]{_{#2,#3}{\hc{(}{#1}\hc{)}}}  %
\newcommand{\sampletimenot}[2]{_{#1}{\hc{(}{#2}\hc{)}}}  %

\newcommand{\graph}{\hc{\mathcal{G}}}
\newcommand{\extendedgraph}{\hc{\tilde{\mathcal{G}}}}
\newcommand{\vertexset}{\hc{\mathcal{V}}}
\newcommand{\extendedvertexset}{\hc{\tilde{\vertexset}}}
\newcommand{\edgeset}{\hc{\mathcal{E}}}
\newcommand{\vertexind}{{\hc{{n}}}}

\newcommand{\vertexindp}{{\hc{{\vertexind}'}}} 
\newcommand{\vertexnum}{{\hc{{N}}}}
 
\newcommand{\featurevecdim}{{\hc{D}} }
 
\newcommand{\adjacencymat}{\hc{\boldmatvec A}} 
\newcommand{\adjacencymatentry}{\hc{ A}} 
\newcommand{\extendedadjacencymat}{\hc{\tilde{\boldmatvec A}}} 
\newcommand{\timeconnectionmat}{\hc{\boldmatvec B}\alongtime} 
\newcommand{\timeconnection}{\hc{b}\alongtime} 
 
\newcommand{\laplacianmat}{\hc{\boldmatvec L}}

\newcommand{\laplacianevecmat}{\hc{\boldmatvec U}} 
\newcommand{\laplacianeval}{\hc{\lambda}} 
 
\newcommand{\alongspace}{ }
\newcommand{\alongtime}{^{(\timeset)}} 
\newcommand{\spaceadjacencymat}{\hc{\boldmatvec A}\alongspace} 
\newcommand{\spaceadjacencymatentry}{\hc{A}}

\newcommand{\signalfun}{{\hc{f}}}
\newcommand{\extendedsignalfun}{{\hc{\tilde f}}}

\newcommand{\signalvec}{\hc{\boldmatvec \signalfun}} 
\newcommand{\extendedsignalvec}{\hc{\boldmatvec{\tilde{ \signalfun}}}}

\newcommand{\fouriersignalfun}{\hc{\fourier f}} 

\newcommand{\signalestfun}{\hc{\hat \signalfun}} 
\newcommand{\signalestvec}{\hc{\hat{\boldmatvec \signalfun}}}

\newcommand{\extendedsignalestvec}{\hc{\hat{\tilde{\boldmatvec \signalfun}}} }
\newcommand{\signalridgeestvec}{\hc{\hbm f}_\text{RR}}

\newcommand{\signallmmseestvec}{\hc{\hbm f}_\text{LMMSE}}

\newcommand{\sampleset}{\hc{\mathcal{S}}} 
\newcommand{\samplemat}{\hc{\boldmatvec S}} 
\newcommand{\extendedsamplemat}{\hc{\tilde{\samplemat}}} 
\newcommand{\sampleind}{{\hc{s}}} 
\newcommand{\samplenum}{{\hc{S}}} 
\newcommand{\samplenumdiagmat}{{\hc{\boldmatvec D}}} 
\newcommand{\extendedsamplenum}{{\hc{\tilde S}}} 
\newcommand{\observationfun}{{\hc{y}}}

\newcommand{\observationvec}{\hc{\boldmatvec y}} 
 
\newcommand{\extendedobservationvec}{\hc{\tilde{\boldmatvec y}}}

\newcommand{\observationnoisefun}{{\hc{e}}} 
\newcommand{\observationoutlyingnoisefun}{\hc{o}}
\newcommand{\observationnoisevec}{{\boldmatvec\observationnoisefun}}

\newcommand{\observationnoisevar}{\hc{ {\sigma^2_\observationnoisefun}}}  %
\newcommand{\paramind}{\hc{m}}
\newcommand{\paramnum}{\hc{M}}
\newcommand{\fp}{\hc{f_\text{P}}}
\newcommand{\fpvec}{\hc{\boldmatvec f}_\text{P}}
\newcommand{\fnp}{\hc{f_\text{NP}}}
\newcommand{\fnpvec}{\hc{\boldmatvec f}_\text{NP}}

\newcommand{\basisfun}{\hc{b}}
\newcommand{\basismat}{\hc{\boldmatvec B}}
\newcommand{\paramcoef}{\hc{\beta}}

\newcommand{\paramcoefvec}{\hc{\boldmatvecgreek {\upbeta}}}
\newcommand{\paramcoefestvec}{\hc{\hat{\boldmatvecgreek{\upbeta}}}}
\newcommand{\samplebasismat}{\hc{\bar{\boldmatvec B}}} 
 
\newcommand{\rkhs}{{\hc{\mathcal{H}}}}
\newcommand{\rkhsfunsymbol}{f}
\newcommand{\rkhsvec}{\hc{{\boldmatvec \rkhsfunsymbol}}}
\newcommand{\extendedrkhsvec}{\hc{{\tilde{ \boldmatvec \rkhsfunsymbol}}}}
\newcommand{\rkhsfun}{\hc{\rkhsfunsymbol}} 
\newcommand{\rkhsestfun}{\hc{\hat \rkhsfunsymbol}} 
 
\newcommand{\fourierrkhsfun}{\hc{\check{ \rkhsfunsymbol}}} 
\newcommand{\fourierrkhsvec}{\hc{\check{\bm \rkhsfunsymbol}}}

\newcommand{\kernelfun}{\hc{\kappa}} 
\newcommand{\fullkernelmat}{\hc{\boldmatvec K}} 
\newcommand{\spacefullkernelmat}{\fullkernelmat\alongspace}

\newcommand{\kernelinvondiagonalmat}{\hc{\boldmatvec D}} 
\newcommand{\kernelinvoffdiagonalmat}{\hc{\boldmatvec C}} 
\newcommand{\extendedfullkernelmat}{\hc{\tilde{ \boldmatvec K}}}

\newcommand{\frequencyweightfun}{\hc{r}}

\newcommand{\samplekernelmat}{\hc{\bar{\boldmatvec K}}} 
\newcommand{\fullalpha}{\hc{ \alpha}} 
\newcommand{\fullalphaestvec}{\hc{\hat{\boldmatvecgreek \upalpha}}} 
\newcommand{\samplealpha}{\hc{\bar \alpha}}
\newcommand{\samplealphaest}{\hc{\hat{\bar \alpha}}} 
\newcommand{\samplealphavec}{\hc{\bar{\boldmatvecgreek \upalpha}}}
\newcommand{\samplealphaestvec}{\hc{\hat{\bar{\boldmatvecgreek \upalpha}}}}

\newcommand{\blnum}{{\hc{B}}} 
\newcommand{\covmat}{\hc{\boldmatvec C}}  %
\newcommand{\rkhsnum}{{\hc{M}}}
\newcommand{\rkhsind}{{\hc{m}}}
\newcommand{\trsamplealphavec}{\hc{\check{\samplealphavec}}} 

\newcommand{\kernelcoef}{{\hc{ \theta}}}
\newcommand{\kernelcoefvec}{{\hc{\boldmatvecgreek \uptheta}}}

\newcommand{\transitionmat}{{\hc{\boldmatvec P}}}
\newcommand{\errormat}{{\hc{\boldmatvec M}}}
\newcommand{\plantnoise}{\eta} 
 
\newcommand{\plantnoisevec}{\hc{\boldmatvecgreek \upeta}} 

\newcommand{\plantnoisekernelmatentry}{{\hc{\Sigma}}} 
\newcommand{\plantnoisekernelmat}{{\hc{\boldmatvecgreek \plantnoisekernelmatentry}}} 
\newcommand{\kalmangainmat}{{\hc{\boldmatvec G}}}

\newcommand{\localset}{\mathcal{V}}

\newcommand{\regpar}{\hc{\mu}}

\newcommand{\fourier}[1]{\check{#1}}

\usepackage{epstopdf}

\usepackage{upgreek}
\usepackage{algorithmic}
\usepackage[ruled,vlined,resetcount]{algorithm2e}
\usepackage{subfigure}
\usepackage{tabularx}
\usepackage{booktabs}
\usepackage{verbatim}   
\usepackage{accents}

\usepackage{pgfplots}
\pgfplotsset{compat=newest}
\usetikzlibrary{plotmarks}
\usetikzlibrary{arrows.meta}
\usepgfplotslibrary{patchplots}
\usepackage{grffile}

\pgfplotsset{plot coordinates/math parser=false}
\newlength\mywidth
\newlength\myheight
\definecolor{mycolor1}{rgb}{0.00000,0.44700,0.74100}%
\definecolor{mycolor2}{rgb}{0.85000,0.32500,0.09800}%
\definecolor{mycolor3}{rgb}{0.92900,0.69400,0.12500}%
\definecolor{mycolor4}{rgb}{0.49400,0.18400,0.55600}%
\definecolor{mycolor5}{rgb}{0.46600,0.67400,0.18800}%
\definecolor{mycolor6}{rgb}{0.30100,0.74500,0.93300}%
\definecolor{mycolor7}{rgb}{0.63500,0.07800,0.18400}%

\definecolor{colorKKrKF}{rgb}{0.00000,0.44700,0.74100}%
\definecolor{colorMKrKF}{rgb}{1,0.0032500,0.001}%
\definecolor{colorLMS}{rgb}{0.92900,0.69400,0.12500}%
\definecolor{colorLMS2}{rgb}{0.49400,0.18400,0.55600}%
\definecolor{colorLMS3}{rgb}{0.49400,0.68400,0.55600}%
\definecolor{colorLMS4}{rgb}{0.49400,0.68400,0.95600}%
\definecolor{colorDLSR}{rgb}{0.46600,0.67400,0.18800}%
\definecolor{colorDLSR2}{rgb}{0.30100,0.74500,0.93300}%
\definecolor{colorDLSR3}{rgb}{0.63500,0.07800,0.18400}%

\theoremstyle{definition}
\theoremstyle{plain}

\newcommand{\cmt}[1]{} 
\newcommand{\hc}[1]{\textcolor{black}{#1}} 

\begin{document}

\title{Kernel-based Inference of Functions over Graphs} %

\author {Vassilis N. Ioannidis$^{\star}$,  Meng Ma$^{\star}$, Athanasios N. 
Nikolakopoulos$^{\star}$, \\ Georgios
	B. Giannakis$^{\star}$, and Daniel Romero$^\dagger$
	\thanks{		
		$^{\star}$
		ECE Dept. and the Digital Tech. Center,
		Univ. of Minnesota, Mpls, MN 55455, USA.
		E-mails:\{ioann006,maxxx971,anikolak,georgios\}@umn.edu, and
	$^\dagger$
	Dept. of Information and Communication Technology
	University of Agder, Grimstad, Norway
	E-mail: daniel.romero@uia.no}
	}
\maketitle
{\footnotesize\begin{center}
		To be published as a chapter in `\textbf{Adaptive Learning Methods for 
			Nonlinear System Modeling}', Elsevier Publishing, Eds. D. 
			Comminiello 
		and J.C. Principe (2018)\end{center}}



\begin{abstract}
The study of networks 
has witnessed an explosive growth over the past decades with several ground-breaking methods introduced.
 A particularly interesting -- and prevalent in several fields of study -- problem is that of \textit{inferring a function defined over the nodes of  a network}.  This work presents a versatile  kernel-based framework for tackling this inference problem that naturally  subsumes and generalizes the reconstruction approaches put forth recently by the signal processing on graphs community. 
 Both the \textit{static} and the \textit{dynamic} settings are considered along with effective modeling approaches for addressing   real-world problems. The herein analytical discussion is complemented by a set of numerical examples, which showcase the effectiveness of the presented techniques, as well as their  merits related to state-of-the-art methods.
\end{abstract}



\section{Introduction}
\label{sec:intro}

\definecolor{mycolor1}{rgb}{0.00000,0.44700,0.74100}%
\definecolor{mycolor2}{rgb}{0.85000,0.32500,0.09800}%
\definecolor{mycolor3}{rgb}{0.92900,0.69400,0.12500}%
\definecolor{mycolor4}{rgb}{0.49400,0.18400,0.55600}%
\definecolor{mycolor5}{rgb}{0.46600,0.67400,0.18800}%
\definecolor{mycolor6}{rgb}{0.30100,0.74500,0.93300}%
\definecolor{mycolor7}{rgb}{0.63500,0.07800,0.18400}%

	Numerous applications arising in diverse disciplines involve \textit{inference over
	networks}~\cite{kolaczyck2009}. 
	 Modelling nodal attributes as signals that take values over the vertices of the underlying graph, allows the associated inference tasks to leverage node dependencies captured by the graph structure.  
 In many real settings one often affords to work with only a 
	limited number of node 
	observations due to 
	inherent restrictions particular to the inference task at hand. In social 
	networks, for example, individuals may be reluctant to share personal 
	information; 
	in sensor networks the nodes may report observations sporadically in order 
	to save energy; 
	in brain networks acquiring 
	node samples 
	may involve invasive procedures (e.g. 
	electrocorticography). 
	In this context, a frequently encountered challenge that often emerges is that of  
 	\textit{inferring the attributes for every node in the network given the attributes for a subset of nodes}.  This is typically formulated as the task of reconstructing a function defined on the  nodes~\cite{kolaczyck2009,kondor2002diffusion,smola2003kernels,shuman2013emerging,sandryhaila2013discrete,chapelle2006},
	 given information about some of its values.

Reconstruction of functions over graphs has been studied by the machine learning community, in the context of \emph{semi-supervised
	learning} under the term of \emph{transductive} regression
and
classification~\cite{chapelle2006,chapelle1999transductive,cortes2007transductive}. 
Existing approaches assume 
``smoothness'' with respect to the graph -- in the sense that neighboring vertices have similar values -- and devise    
\emph{nonparametric}
methods~\cite{kondor2002diffusion,smola2003kernels,chapelle2006,belkin2006manifold} targeting primarily the task of reconstructing binary-valued signals.  
Function estimation has also been investigated
recently by the community of signal processing on graphs (SPoG)
under the term \emph{signal
	reconstruction}~\cite{narang2013structured,narang2013localized,gadde2015probabilistic,tsitsvero2016uncertainty,chen2015theory,anis2016proxies,wang2015local,marques2015aggregations}.
 Most such approaches commonly adopt \emph{parametric}
estimation tools, and rely on \emph{bandlimitedness}, by
which the signal of interest is assumed to lie in the span of the
$\blnum$ principal eigenvectors of the graph's Laplacian (or adjacency) matrix.

This chapter cross-pollinates ideas arising from both communities, and presents a unifying framework for tackling signal reconstruction problems both in the traditional \textit{time-invariant}, as well as in the more challenging \textit{time-varying} setting.
We begin by a comprehensive presentation of kernel-based learning for solving problems of signal reconstruction over graphs (Section~\ref{sec:tirecon}). 
Data-driven techniques are then presented based on multi-kernel learning (MKL) that enables combining optimally the kernels in a given
dictionary, and simultaneously estimating the graph function by solving a
single optimization problem 
(Section~\ref{sec:select_kernels}).  For the case where prior information is available, semi-parametric estimators are discussed  that can incorporate seamlessly structured prior information into the signal estimators (Section~\ref{sec:semipar}). 
We then move to the problem of reconstructing time-evolving functions on dynamic graphs (Section~\ref{sec:timevar}). The kernel-based framework   
is now extended to accommodate the time-evolving setting building on  the notion of \emph{graph extension}, specific choices of which can lend themselves to a reduced complexity online solver (Section~\ref{sec:reconstruction}). Next, a more flexible model is introduced that captures multiple forms of time dynamics, and  kernel-based learning is employed to derive an online solver, that effects online MKL by selecting the optimal combination of kernels on-the-fly (Section~\ref{sec:multkr}). Our analytical exposition, in both parts, is supplemented by a set of numerical tests based on both real and synthetic data that highlight the effectiveness of the methods, while providing examples of interesting realistic problems that they can address.

\noindent{\textbf{Notation}:}
 Scalars are denoted by
lowercase characters, vectors by bold lowercase,  matrices by bold
uppercase;  
   $( A)_{m,n}$ is the  $(m,n)$-th
  entry of matrix $\boldmatvec A$; 
   superscripts $~\transpose$  and   $~\pinv$  respectively
  denote transpose and pseudo-inverse.
   If $\boldmatvec A\define[\boldmatvec a_1,\ldots,\boldmatvec a_N]$, then 
   $\vect\{\boldmatvec
  A\}
\define [\boldmatvec a_1\transpose,\ldots,\boldmatvec 
a_N\transpose]\transpose\define\boldmatvec a$.
   With $\vertexnum\times\vertexnum$ matrices $\{\boldmatvec 
   A_\timeind\}_{\timeind=1}^\timenum$
  and $\{\boldmatvec B_\timeind\}_{\timeind=2}^\timenum$ satisfying 
  $\boldmatvec A_\timeind
  = \boldmatvec A_\timeind\transpose~\forall\timeind$, 
$ \tridiag\{\boldmatvec A_1,\ldots,\boldmatvec A_\timenum;\boldmatvec 
B_2,\ldots,\boldmatvec B_\timenum\}$
represents the symmetric block tridiagonal matrix.  Symbols $\odot$, $\otimes$, 
and $\oplus$ 
  respectively denote element-wise (Hadamard) matrix product,
  Kronecker product, and Kronecker sum, the latter being defined for
  $\boldmatvec A\in \mathbb{R}^{M \times M}$ and $\boldmatvec B\in \rfield^{N 
  \times N}$
  as $\boldmatvec A \oplus \boldmatvec B \define \boldmatvec A \otimes 
  \boldmatvec I_N + \boldmatvec I_M
  \otimes \boldmatvec B$.   The $n$-th column of the identity matrix
  $\identitymat_N$ is represented by $\canonicalvec{N}{n}$.  
  If $\boldmatvec A\in \rfield^{N \times N}$ is positive definite and 
  $\boldmatvec
  x\in \rfield^N$, then $||\boldmatvec x||^2_{\boldmatvec A}\define \boldmatvec 
  x\transpose
  \boldmatvec A\inv \boldmatvec x$ and $||\boldmatvec x ||_2\define 
  ||\boldmatvec
  x||_{\identitymat_N}$. The cone of $N\times N$ positive
  definite matrices is denoted by $\pdset^N$. Finally,
  $\delta[\cdot]$ stands for the Kronecker delta, 
  and $\expectednb$ for  expectation.

\section{Reconstruction of Functions over Graphs}
\label{sec:tirecon}

Before giving the formal problem statement, it is instructive to start with the basic definitions that will be used throughout this chapter.

\noindent{\textbf{Definitions}:} A graph can be specified by a tuple 
$\mathcal{G}:=(\vertexset,\adjacencymat)$, where
$\vertexset:=\{v_1, \ldots, v_N\}$ is the vertex set, and $\adjacencymat$ is the
$\vertexnum\times \vertexnum$ adjacency matrix, whose
$(\vertexind,\vertexindp)$-th entry, 
$\adjacencymatentry\vertexvertexnot{\vertexind}{\vertexindp}\ge0$, denotes the non-negative edge 
weight between vertices 
$v_\vertexind$ and
$v_{\vertexindp}$.  
For simplicity, it is assumed
that the graph has no self-loops, i.e. 
$\adjacencymatentry\vertexvertexnot{\vertexind}{\vertexind}=0,~\forall v_\vertexind\in 
\vertexset$.  This chapter focuses on
\emph{undirected} graphs, for which $\adjacencymatentry\vertexvertexnot{\vertexindp}{\vertexind} = 
\adjacencymatentry\vertexvertexnot{\vertexind}{\vertexindp}~\forall 
v_\vertexind,v_\vertexindp\in
\vertexset$.  A graph is said to be \emph{unweighted} if 
$\adjacencymatentry\vertexvertexnot{\vertexind}{\vertexindp}$ is
either 0 or 1.  The edge set is defined as $\edgeset :=\{(v_\vertexind,v_\vertexindp)\in 
\vertexset \times 
\vertexset: \adjacencymatentry\vertexvertexnot{\vertexind}{\vertexindp}\neq 0\}$.  Two vertices 
$v_\vertexind$ 
and $v_\vertexindp$ are \emph{adjacent},
\emph{connected}, or \emph{neighbors} if $(v_\vertexind,v_\vertexindp)\in \edgeset$.  
The \emph{Laplacian} matrix is defined as 
$\boldmatvec{L}:=\diag{\adjacencymat\boldmatvec{1}}-\adjacencymat$, and is 
symmetric and positive
semidefinite~\cite[Ch. 2]{kolaczyck2009}. 
 A real-valued function (or signal) on a graph is a map
$ \signalfun:\mathcal{V} \rightarrow \mathbb{R}$. The value $\signalfun(v)$ represents an
attribute or feature of $v\in \vertexset$, such as age, political
alignment, or annual income of a person in a social network. Signal
$\signalfun$ is thus represented by
$\signalvec:=[\signalfun(v_1),\ldots,\signalfun(v_N)]^T$.

\noindent\textbf{Problem statement.}
\cmt{observations}Suppose that a collection of noisy samples (or
observations) $\{ y_s | y_s = \signalfun(v_{\vertexind_s}) + \noisesamp_s \}_{s=1}^S$ is available, 
where $\noisesamp_s$ models noise
and $\mathcal{S}:=\{\vertexind_1,\ldots,\vertexind_S\}$ contains the
indices $1\leq \vertexind_1<\cdots<\vertexind_S\leq \vertexnum$ of the
sampled vertices, with $S\le\vertexnum$.   
 Given
$\{(\vertexind_\sampleind,y_\sampleind)\}_{\sampleind=1}^\samplenum$,
and assuming knowledge of $\mathcal{G}$, the goal is to estimate
$\signalfun$. This will provide estimates of $\signalfun(v)$ both at
observed and unobserved vertices.
 By defining $\observationvec \define
[y_{1},\ldots,y_{S}]^T$, the observation model is summarized as
\begin{equation}
	\label{eq:observationsvec}
	\observationvec =\samplingmat \signalvec  + \noisevec
\end{equation}
where $\noisevec \define [\noisesamp_{1},\ldots,\noisesamp_{S}]^T$ and $\samplingmat$ is
a known $S\times N$ binary sampling matrix with entries $(s,\vertexind_s)$,
$s=1,\ldots,S$, set to one, and the rest set to zero.

\subsection{Kernel Regression} Kernel methods constitute the
``workhorse'' of machine learning for nonlinear function
estimation~\cite{scholkopf2002}.  Their popularity can be attributed  to
their simplicity, flexibility, and good performance.  Here, we present kernel regression as a
unifying framework for graph signal reconstruction
along with the so-called representer theorem. 

 Kernel regression seeks an estimate of
  $\signalfun$ in a reproducing kernel Hilbert space (RKHS) $\rkhs$, which is the space of functions
  $\rkhsfun:\vertexset \rightarrow \rfield$ defined as
  \begin{align}
	\label{eq:rkhsdef}
	\rkhs:=\left\{\rkhsfun:\rkhsfun(v) =
	\sum_{\vertexind=1}^{\vertexnum}\fullalpha_\vertexind
	\kernelmap(v,v_\vertexind),~\fullalpha_\vertexind\in \rfield \right\}
\end{align}
where the \emph{kernel map} $\kernelmap:\vertexset\times
\vertexset\rightarrow \rfield$ is any function defining a symmetric
and positive semidefinite $N\times N$ matrix with entries
$[\fullkernelmat]_{\vertexind,\vertexindp} \define
\kernelmap(v_\vertexind,v_\vertexindp)$~\cite{scholkopf2001representer}.  
 Intuitively, $\kernelmap(v,v')$ is a basis function in \eqref{eq:rkhsdef} measuring 
similarity between the values of
$\signalfun$ at $v$ and $v'$. (For a more detailed treatment of RKHS, see e.g.  \cite{smola2003kernels}).

Note that for signals over graphs, the expansion in \eqref{eq:rkhsdef} is finite since $\vertexset$ is finite-dimensional. 
\cmt{General form $\rkhsfun$ in graph RKHS} Thus, any 
$f\in\rkhs$ can be expressed in compact form
\begin{align}
	\label{eq:generalform}
	\rkhsvec 
	= \fullkernelmat  \fullalphavec
\end{align}
for some $N\times 1$ vector $\fullalphavec\define[\fullalpha_1,\ldots,\fullalpha_\vertexnum]^T$.

Given two functions $\rkhsfun(v) \define
  \sum_{\vertexind=1}^{\vertexnum}\fullalpha_\vertexind
  \kernelmap(v,v_\vertexind)$ and $\rkhsfun'(v) \define
  \sum_{\vertexind=1}^{\vertexnum}\fullalpha'_\vertexind
  \kernelmap(v,v_\vertexind)$, their RKHS inner product is defined
  as\footnote{While $\rkhsfun$ denotes a \emph{function}, 
    $\rkhsfun(v)$ represents the \emph{scalar} resulting from
    evaluating $\rkhsfun$ at vertex  $v$. }
\begin{align}
	\label{eq:innerproduct}
	\langle \rkhsfun,\rkhsfun'\rangle_\mathcal{H} := 
	\sum_{\vertexind=1}^{N}\sum_{\vertexindp=1}^{N}\fullalpha_\vertexind \fullalpha'_\vertexindp 
	\kernelmap(
	v_\vertexind,v_\vertexindp) = \fullalphavec^T \fullkernelmat \fullalphavec'
\end{align}
where
$\fullalphavec'\define[\fullalpha'_1,\ldots,\fullalpha'_\vertexnum]^T$ and the reproducing property has been employed that
suggests $\langle
\kernelmap(\cdot,v_{\vertexind_0}),\kernelmap(\cdot,v_{\vertexind_0'})\rangle_\mathcal{H}
= \boldmatvec i_{\vertexind_0}^T \fullkernelmat \boldmatvec i_{\vertexind_0'} =
\kernelmap(v_{\vertexind_0},v_{\vertexind_0'})$. 
 The RKHS norm is defined by
\begin{align}
	\label{eq:norm}
	||\rkhsfun||_\mathcal{H}^2\define \langle \rkhsfun,\rkhsfun \rangle_\rkhs=
	\fullalphavec^T \fullkernelmat \fullalphavec
\end{align}
and will be used as a regularizer to control overfitting and to cope with the under-determined reconstruction problem.
As a special case, setting $\fullkernelmat=\boldmatvec I_N$ recovers the 
standard inner 
product $\langle \rkhsfun,\rkhsfun'\rangle_\rkhs = \rkhsvec^T \rkhsvec'$, and the Euclidean norm 
$||\rkhsfun||_\rkhs^2 = ||\rkhsvec||_2^2$. Note that when $\fullkernelmat\succ 
\boldmatvec 0$, the set of 
functions of the form \eqref{eq:generalform} coincides with $\rfield^\vertexnum$.

 Given $\{y_\sampleind\}_{\sampleind=1}^\samplenum$,
RKHS-based function estimators are obtained by solving functional
minimization problems formulated as (see also
e.g.~\cite{scholkopf2002,scholkopf2001representer,vapnik1998})
\begin{equation}
	\signalestfun := \argmin_{\rkhsfun\in \mathcal{H}} 
	\mathcal{L} (\boldmatvec y, \samplerkhsvec)
	+ \regpar \Omega(||\rkhsfun||_\mathcal{H})
	\label{eq:single-general}
\end{equation}
where the loss $\mathcal{L}$ measures
how the estimated function $\rkhsfun$ at the observed
vertices $\{v_{\vertexind_\sampleind}\}_{\sampleind=1}^\samplenum$, collected in $\samplerkhsvec \define
[\rkhsfun(v_{\vertexind_1}),\ldots,\rkhsfun(v_{\vertexind_\samplenum})]^T=\samplingmat
\rkhsvec$,
deviates from the data $\observationvec$. The so-called \emph{square loss}
$\mathcal{L}(\boldmatvec 
y,\samplerkhsvec)\define{(1/\samplenum)}\sum_{\sampleind=1}^\samplenum\left[y_s-\rkhsfun(v_{\vertexind_s})\right]^2$
constitutes a popular choice for $\mathcal{L}$.  The increasing
function $\Omega$ is used to promote smoothness with typical choices
including $\Omega(\zeta) = |\zeta|$ and $ \Omega(\zeta) = \zeta^2$.  The
regularization parameter $\regpar>0$ controls overfitting. 
\cmt{solution}
Substituting \eqref{eq:generalform} and \eqref{eq:norm} into
\eqref{eq:single-general} shows that $\signalestvec$ can be found as
\begin{subequations}
\begin{align}
   \fullalphaestvec&\displaystyle := \argmin_{\fullalphavec\in \rfield^\vertexnum} 
			\mathcal{L} (\observationvec, \samplingmat \fullkernelmat \fullalphavec)
			+ \regpar \Omega((\fullalphavec^T \fullkernelmat \fullalphavec)^{1/2})\label{eq:alphaopt}\\
			\signalestvec& = \fullkernelmat\fullalphaestvec.
	\label{eq:signalfromalpha}
\end{align}
	\label{eq:solutiongraphs}
\end{subequations}
\cmt{Alternative form}An alternative form of
\eqref{eq:norm} that will be frequently used in the sequel
results upon noting that \begin{align}
	\fullalphavec^T \fullkernelmat \fullalphavec
	= \fullalphavec^T \fullkernelmat \fullkernelmat^\dagger \fullkernelmat
	\fullalphavec = \rkhsvec^T \fullkernelmat^\dagger \rkhsvec.\label{eq:rkhsnormalt}
\end{align} Thus, one
can rewrite \eqref{eq:single-general} as
\begin{equation}
	\signalestvec := \argmin_{\rkhsvec\in \columnspan\{\fullkernelmat\}} 
	\mathcal{L} ( \observationvec,\samplingmat\rkhsvec)
	+ \regpar \Omega((\rkhsvec^T \fullkernelmat^\dagger \rkhsvec)^{1/2}).
	\label{eq:solutiongraphsf}
\end{equation}

Although graph signals can be
reconstructed from \eqref{eq:solutiongraphs}, such an approach
involves optimizing over $N$ variables. 
Thankfully, the 
solution can be obtained by solving an optimization problem in $S$
variables (where typically $S\ll N$), by invoking the so-called representer
theorem~\cite{scholkopf2001representer,kimeldorf1971spline}.

The representer
theorem plays an
instrumental role in the traditional  infinite-dimensional
setting where \eqref{eq:single-general} cannot be solved directly;  
however, even when 
$\rkhs$ comprises graph signals, 
it can still be beneficial to reduce the dimension of
the optimization in \eqref{eq:solutiongraphs}. 
The theorem essentially asserts that the solution to the functional
minimization in \eqref{eq:single-general} can be expressed as
\begin{align}
	\label{eq:representer}
	\signalestfun(v) = \sum_{s=1}^{S} \samplealpha_s \kernelfun(v ,v_{\vertexind_s}) 
\end{align}
for some $\samplealpha_s \in \rfield$, $s=1,\ldots,S$.

\cmt{optimal alphas} The representer theorem shows the form of
$\signalestfun$, but does not provide the optimal
$\{\samplealpha_s\}_{s=1}^S$, which are found after substituting
\eqref{eq:representer} into \eqref{eq:single-general}, and solving the
resulting optimization problem with respect to these coefficients. To
this end, let
$\samplealphavec:=[\samplealpha_1,\ldots,\samplealpha_\samplenum]^T$,
and write  $\fullalphavec = \samplingmat^T \samplealphavec $
to deduce that 
\begin{align}
	\label{eq:graphestimate}
	\signalestvec= \fullkernelmat \fullalphavec
	= \fullkernelmat \samplingmat^T \samplealphavec.
\end{align}
With $\samplekernelmat\define \samplingmat \fullkernelmat \samplingmat^T$ and using \eqref{eq:solutiongraphs} and \eqref{eq:graphestimate}, the
optimal $\samplealphavec$ can be found as
\begin{equation}
	\samplealphaestvec 
	:= \argmin_{\samplealphavec\in \rfield^\samplenum} 
	\mathcal{L} (  \observationvec , \samplekernelmat \samplealphavec)
	+ \regpar \Omega((\samplealphavec^T \samplekernelmat \samplealphavec)^{1/2})
	\label{eq:solutiongraphsmatrixform}.
\end{equation}
\noindent\textbf{Kernel ridge regression}.
For $\mathcal{L}$ chosen as the square loss and
$\Omega(\zeta) =\zeta^2$, the $\signalestfun$ in
\eqref{eq:single-general} is referred to as the \emph{kernel ridge
  regression} (RR) estimate \cite{scholkopf2002}. If $
\samplekernelmat $ is full rank, this estimate is given by
$\signalridgeestvec = \fullkernelmat \samplingmat^T \samplealphaestvec
$, where
\begin{subequations}
	\label{eq:single-lq-m}
\begin{align}
	\label{eq:single-lq-m-a}
	\samplealphaestvec &:= \argmin_{\samplealphavec\in \rfield^\samplenum} 
	\frac{1}{S} \left\| \boldmatvec{y} -
\samplekernelmat\boldmatvec{\samplealphavec} \right\|^2 + \regpar 
\boldmatvec{\samplealphavec}^T \samplekernelmat
\boldmatvec{\samplealphavec}\\
&= (\samplekernelmat + \regpar S \boldmatvec{I}_S )^{-1} \boldmatvec{y} .
\end{align}
\end{subequations}
Therefore, $\signalridgeestvec$ can be expressed as
\begin{align}
\label{eq:ridgeregressionestimate}
\signalridgeestvec
= \fullkernelmat \samplingmat^T (\samplekernelmat + \regpar S \boldmatvec{I}_S 
)^{-1} \boldmatvec{y}.
\end{align}
As we will see in the next section,
\eqref{eq:ridgeregressionestimate} generalizes a number of existing
signal reconstructors upon properly selecting $\fullkernelmat$. 

\subsection{Kernels on Graphs}
\label{sec:kernelsong}
%
%
When estimating functions on graphs, conventional
kernels such as the Gaussian kernel cannot be adopted because the underlying
set where graph signals are defined is not a metric space. Indeed, no
vertex addition $v_{\vertexind} + v_{\vertexind'}$, scaling $\beta
v_{\vertexind}$, or norm $||v_{\vertexind}||$ can be naturally defined
on $\vertexset$. An alternative is to embed
$\vertexset$ into Euclidean space via a feature {map}
$\phi:\vertexset \rightarrow \rfield^\featurevecdim$, and invoke a
conventional kernel afterwards.  However, for a given graph it is
generally unclear how to explicitly  design $\phi$  or select
$\featurevecdim$. This motivates the adoption of \textit{kernels on graphs}~\cite{smola2003kernels}.

A common approach to designing kernels on graphs is to apply a transformation function
on the graph Laplacian~\cite{smola2003kernels}.
 The term \textit{Laplacian kernel} comprises a wide family of
kernels obtained by applying a certain function $r(\cdot)$ to the Laplacian matrix
$\boldmatvec L$. Laplacian kernels are well
motivated since they constitute the graph counterpart of the so-called
\emph{translation-invariant kernels} in Euclidean
spaces~\cite{smola2003kernels}.  This section
reviews Laplacian kernels, provides beneficial insights in terms of
interpolating signals, and highlights their versatility in capturing
information about the \emph{graph Fourier transform} of the
estimated signal.

The reason why the graph Laplacian 
constitutes one of the prominent candidates for regularization on graphs, 
becomes clear upon recognizing that 
\begin{equation}
	\label{eq:graphvariation}
	\rkhsvec^T \laplacianmat \rkhsvec = \frac{1}{2} \sum_{(n,n') \in \edgeset} A_{n,n'} (\rkhsfun_n - \rkhsfun_{n'})^2,
\end{equation}
where $ A_{n,n'} $ denotes weight associated with edge $ (n,n') $. The quadratic form of~\eqref{eq:graphvariation} becomes larger when function values vary a lot among connected vertices and therefore quantifies the \emph{smoothness} of $\rkhsvec$ on $\graph$.
 
Let $0=\lambda_1\leq \lambda_2\leq \ldots\leq
\lambda_N$ denote the eigenvalues of the graph Laplacian matrix
$\boldmatvec L$, and consider the eigendecomposition $\boldmatvec
L=\laplacianevecmat \boldmatvec \Lambda \laplacianevecmat^T$, where $\boldmatvec
\Lambda\define\diag{\lambda_1,\ldots,\lambda_N}$. A Laplacian kernel matrix is defined by 
\begin{align}
	\fullkernelmat\define r^\dagger(\boldmatvec L) \define \laplacianevecmat
	r^\dagger(\boldmatvec \Lambda) \laplacianevecmat^T
	\label{eq:laplacian_kernel}
\end{align}
where $r(\boldmatvec \Lambda)$ is the result of applying a user-selected, 
scalar, 
non-negative map $r:\rfield\rightarrow \rfield_+$ to the diagonal
entries of $\boldmatvec \Lambda$.  The selection of map $ r $ generally depends 
on desirable properties that the target function is expected to have. 
Table~\ref{tab:spectralweightfuns} summarizes some well-known examples
arising for specific choices of $\frequencyweightfun$.

\begin{table}
	\begin{center}
		\begin{tabular}{p{5cm}  l  p{2 cm}}
			\bottomrule
			\textbf{Kernel name}  & \textbf{Function}  & \textbf{Parameters} \\
			\bottomrule
			Diffusion~\cite{kondor2002diffusion}     &
			$r(\lambda)=\exp\{\sigma^2\lambda/2\}$  & 
			$\sigma^2$
			\\
			\hline
			$p$-step random walk~\cite{smola2003kernels}    & $r(\lambda) =
			(a-\lambda)^{-p}$ & $a\geq 2$, $p\ge 0$  \\
			\hline
			Regularized Laplacian~\cite{smola2003kernels,zhou2004regularization}
			& $r(\lambda)=1 + \sigma^2\lambda$  & $\sigma^2$ \\
			\hline
			Bandlimited~\cite{romero2017kernel}    &  
			$\begin{aligned}
			\label{eq:defrbl}
			\frequencyweightfun(\laplacianeval) =  
			\begin{cases}
			1/\beta & \laplacianeval\leq\laplacianeval_\text{max}\\
			\beta & \text{otherwise}
			\end{cases}
			\end{aligned}$
			&
			$\beta>0$,   $\lambda_\text{max}$
			\\
			\bottomrule
		\end{tabular}
	\end{center}
	\caption{Common spectral weight functions.}
	\label{tab:spectralweightfuns}
\end{table}

At this point, it is prudent to offer interpretations and insights on
the operation of Laplacian kernels.  \cmt{rewriting
  regularizer}Towards this objective, note first that the regularizer
from~\eqref{eq:solutiongraphsf} is an increasing function of
\begin{align}
	\fullalphavec^T \fullkernelmat \fullalphavec
	= \fullalphavec^T \fullkernelmat \fullkernelmat^\dagger \fullkernelmat
	\fullalphavec = \rkhsvec^T  \fullkernelmat^\dagger   \rkhsvec
	= \rkhsvec^T \laplacianevecmat r(\boldmatvec \Lambda)
	\laplacianevecmat^T \rkhsvec
	= \fourierrkhsvec^T  r(\boldmatvec \Lambda)
	\fourierrkhsvec =
	\sum_{\vertexind=1}^Nr(\lambda_\vertexind)|\fourierrkhsfun_\vertexind|^2
	\label{eq:fourierregularizer}
\end{align}
where $\fourierrkhsvec \define \laplacianevecmat^T \rkhsvec\define
[\fourierrkhsfun_1,\ldots,\fourierrkhsfun_N]^T $ comprises the
projections of $\rkhsvec$ onto the eigenspace of $\boldmatvec L$, and is
referred to as the \emph{graph Fourier transform} of $\rkhsvec$ in the
SPoG parlance~\cite{shuman2013emerging}. Consequently, $ \{\fourierrkhsfun_n\}_{n=1}^{N} $ are called \emph{frequency components}. \cmt{bandlimited function} The so-called \emph{bandlimited functions} in SPoG refer to those whose frequency components only exist inside some band $ B $, that is, $ \fourierrkhsfun_n = 0, \forall n > B $.

By adopting the
aforementioned SPoG notions, one can intuitively interpret the role
of bandlimited kernels. Indeed, it follows from
\eqref{eq:fourierregularizer} that the regularizer strongly penalizes those
$\fourierrkhsfun_\vertexind$ for which the corresponding
$r(\lambda_\vertexind)$ is large, thus promoting a specific structure
in this ``frequency'' domain. Specifically, one prefers
$r(\lambda_\vertexind)$ to be large whenever $|\fourierrkhsfun_n|^2$
is small and vice versa. The fact that
$|\fourierrkhsfun_\vertexind|^2$ is expected to decrease with
$\vertexind$ for smooth $\rkhsfun$, motivates the adoption of an
increasing function $r$~\cite{smola2003kernels}. From~\eqref{eq:fourierregularizer} it is 
clear that $ r(\lambda_\vertexind) $ determines how heavily $ 
\fourierrkhsfun_\vertexind $ is penalized. Therefore, by setting $ 
r(\lambda_\vertexind) $ to be small when $ \vertexind \leq B $ and extremely 
large when $ \vertexind > B $, one can expect the result to be a bandlimited 
signal.

Observe that 
Laplacian kernels
can capture forms of prior information richer than 
bandlimitedness~\cite{narang2013localized,tsitsvero2016uncertainty,wang2015local,marques2015aggregations}
by selecting function $r$ accordingly. 
For instance, using $r(\lambda)=\exp\{\sigma^2\lambda/2\}$ (diffusion kernel) accounts not only for smoothness of $\mathbf{f}$ as in~(\ref{eq:graphvariation}), but also for the prior that $\mathbf{f}$ is generated by a  process diffusing over the graph. Similarly, the use of $r(\lambda)=(\alpha - \lambda)^{-1}$ (1-step random walk) can accommodate cases where the signal captures a notion of network centrality\footnote{Smola et al.~\cite{smola2003kernels}, for example, discuss the connection between $r(\lambda)=(\alpha - \lambda)^{-1}$ and PageRank~\cite{page1999pagerank} whereby the sought-after signal is essentially defined as the limiting distribution of a simple underlying ``\textit{random surfing}" process. For more about random surfing processes, see also ~\cite{nikolakopoulosBTRank,nikolakopoulos2015random}.}.

So far, $\signalfun$ has been assumed
deterministic, which precludes accommodating certain forms of prior
information that probabilistic models can capture, such as domain
knowledge and historical data. 
Suppose without loss of generality that
$\{\signalfun(v_\vertexind)\}_{\vertexind=1}^{\vertexnum}$ are
zero-mean random variables. The LMMSE estimator
of $\signalvec$ given $\observationvec$ in (\ref{eq:observationsvec}) is the linear estimator
$\signallmmseestvec$ minimizing $\expectednb||\signalvec-
\signallmmseestvec ||_2^2$, where the expectation is over all
$\signalvec$ and noise realizations. With $\covmat
\define\expected{\signalvec \signalvec^T}$, the LMMSE estimate is 
\begin{align}
\label{eq:lmmseestimator}
\signallmmseestvec = \covmat \samplingmat^T[\samplingmat \covmat
\samplingmat^T + \noisevar \boldmatvec I_S]\inv \observationvec
\end{align}
where $\noisevar\define ({1}/{S})\expected{||\noisevec||_2^2}$ denotes
the noise variance. Comparing
\eqref{eq:lmmseestimator} with \eqref{eq:ridgeregressionestimate} and
recalling that $\samplekernelmat\define \samplingmat \fullkernelmat
\samplingmat^T$, it follows that $\signallmmseestvec =
\signalridgeestvec$ if $\mu S = \noisevar$ and
$\fullkernelmat = \covmat$. In other words, the similarity measure
$\kernelfun(v_\vertexind,v_\vertexindp)$ embodied in such a kernel map is
just the covariance $\mathop{\rm
  cov}[\signalfun(v_\vertexind),\signalfun(v_\vertexindp)]$. A related
observation was pointed out in~\cite{bazerque2013basispursuit} for
general kernel methods.

In short, one can
interpret kernel ridge regression as the LMMSE estimator of a signal
$\signalvec$ with covariance matrix equal to $\fullkernelmat$;
see also \cite{sindhwani2005transductivesemisupervised}. 
  The LMMSE interpretation also suggests the
  usage of $\covmat$ as a kernel matrix, which enables signal
  reconstruction even when the graph topology is unknown. Although
  this discussion hinges on kernel ridge regression after setting
  $\fullkernelmat =\covmat$, any other kernel estimator of the form
  \eqref{eq:solutiongraphs} can benefit from vertex-covariance kernels too.

 In most contemporary  
 networks, 	sets of nodes may depend among each other via multiple types of 
 relationships, which ordinary networks cannot 
 capture~\cite{kivela2014multilayer}. Consequently, generalizing the 
 traditional \emph{single-layer} to \emph{multilayer} 
 networks that organize the nodes into different groups, called 	
 \emph{layers},is well motivated. For kernel-based approaches for function reconstruction over multilayer graphs see also \cite{ioannidis2018multilay}.
\subsection{Selecting kernels from a dictionary}
\label{sec:select_kernels}

 The selection of the  pertinent kernel matrix is of paramount importance to the 
performance of kernel-based methods~\cite{romero2017kernel,gonen2011multiple}. 
This section presents an MKL approach that effects
kernel selection in graph signal reconstruction. Two
algorithms with complementary strengths will be presented. Both rely on a user-specified \emph{kernel 
	dictionary}, and the best kernel is built from the dictionary in a data driven way.

The first algorithm, which we call \emph{RKHS superposition}, is motivated 
by the fact that one specific $\mathcal{H}$ in
\eqref{eq:single-general} is determined by some $\kernelmap$; therefore, kernel selection is 
tantamount to RKHS selection. Consequently, a kernel dictionary 
$\{\kernelmap_m\}_{\rkhsind=1}^\rkhsnum$ gives rise to a 
RKHS dictionary $\{\mathcal{H}_m\}_{\rkhsind=1}^\rkhsnum$, which
motivates estimates of the form\footnote{A sum is chosen here
	for tractability, but the right-hand side of \eqref{eq:sumh} could
	in principle combine the functions $\{\rkhsestfun_m\}_m$ in
	different forms. }
\begin{equation}
	\label{eq:sumh}
	{\rkhsestfun} = \sum_{m=1}^\rkhsnum {\rkhsestfun}_m, \quad {\rkhsestfun}_m \in \mathcal{H}_m.
\end{equation}
Upon adopting a criterion that controls sparsity in
this expansion, the ``best'' RKHSs will be selected. A reasonable
approach is therefore to generalize \eqref{eq:single-general} to
accommodate multiple RKHSs. With $\mathcal{L}$ selected to be the square
loss and $\Omega(\zeta)=|\zeta|$, one can pursue an estimate
$\rkhsestfun$ by solving
\begin{equation}
	\hspace{-.25cm}\min_{\{\rkhsfun_\rkhsind\in \mathcal{H}_m\}_{m=1}^M} \frac{1}{S} \sum_{s=1}^S 
	\left[y_s - \sum_{m=1}^M
	\rkhsfun_\rkhsind(v_{\vertexind_s}) \right]^2 + \regpar 
	\sum_{m=1}^M\|\rkhsfun_\rkhsind\|_{\mathcal{H}_m}.
	\label{eq:multi-sq}
\end{equation}
Invoking the representer theorem
per $\rkhsfun_\rkhsind$ establishes that the minimizers of
\eqref{eq:multi-sq} can be written as
\begin{align}
	\label{eq:representer-mkl-g}
	\hat \rkhsfun_\rkhsind(v) = \sum_{s=1}^{S} \samplealpha^m_s \kernelmap_m(v,v_{\vertexind_s}) 
	,~~~~m=1,\ldots,M
\end{align}
for some coefficients $\samplealpha^m_s$. Substituting
\eqref{eq:representer-mkl-g} into \eqref{eq:multi-sq} suggests
obtaining these coefficients as
\begin{equation}
	\argmin_{\{\samplealphavec_\rkhsind\}_{m=1}^M} \frac{1}{S} \left\| 
	\boldmatvec{y}
	- \sum_{m=1}^M\samplekernelmat_m\samplealphavec_\rkhsind \right\|^2 + \regpar \sum_{m=1}^M 
	\left(
	\samplealphavec_\rkhsind^T \samplekernelmat_m \samplealphavec_\rkhsind \right)^{1/2}
	\label{eq:multi-vector-sq}
\end{equation}
where $\samplealphavec_\rkhsind:=[\samplealpha_1^m,\ldots,\samplealpha_S^m]^T$,
and $\samplekernelmat_\rkhsind:=\samplingmat \fullkernelmat_m
\samplingmat^T$ with
$(\fullkernelmat_m)_{\vertexind,\vertexindp}:=\kernelmap_m(v_{\vertexind},v_{\vertexindp})$.
\cmt{promotes sparsity} 
Letting $\trsamplealphavec_m := \samplekernelmat_m^{1/2}\samplealphavec_m$, expression
\eqref{eq:multi-vector-sq} becomes
\begin{align}
	&\argmin_{\{\trsamplealphavec_m\}_{m=1}^M} \frac{1}{S} \left\| 
	\boldmatvec{y} -
	\sum_{m=1}^M\samplekernelmat_m^{1/2}\trsamplealphavec_m \right\|^2 + \regpar \sum_{m=1}^M \| 
	\trsamplealphavec_m
	\|_2 	\label{eq:multi-1norm}.
\end{align}

Interestingly, \eqref{eq:multi-1norm} can be efficiently
solved using the alternating-direction method of multipliers
(ADMM)~\cite{bazerque2011splines,giannakis2016decentralized} after some nessecary reformulation~\cite{romero2017kernel}. 

\cmt{final estimate} After obtaining
${\{\trsamplealphavec_m\}_{m=1}^M}$,
the sought-after function estimate can be recovered as
\begin{align}
	\label{eq:mklrkhsreconstruction}
	\signalestvec = \sum_{m=1}^M\fullkernelmat_m \samplingmat^T
	\samplealphavec_m= \sum_{m=1}^M\fullkernelmat_m \samplingmat^T
	\samplekernelmat_m^{-1/2}\trsamplealphavec_m.
\end{align}

%
%
 This MKL algorithm
can identify the best subset of RKHSs -- and therefore kernels -- but
entails $\rkhsnum\samplenum $ unknowns
(cf. \eqref{eq:multi-vector-sq}). Next, an
alternative approach is discussed which can reduce the number of variables to $\rkhsnum+\samplenum$
 at the price of not beeing able to assure a sparse kernel
expansion.

 The alternative approach is to postulate a kernel 
of the form
$ \fullkernelmat(\kernelcoefvec) = \sum_{\rkhsind=1}^{\rkhsnum}
\kernelcoef_\rkhsind \fullkernelmat_\rkhsind $, where
$\{\fullkernelmat_\rkhsind\}_{\rkhsind=1}^\rkhsnum$ is given and $
\kernelcoef_\rkhsind\geq 0~\forall \rkhsind$. The coefficients
$\kernelcoefvec\define[\kernelcoef_1,\ldots,\kernelcoef_\rkhsnum]^T$
can be found by jointly minimizing
\eqref{eq:solutiongraphsmatrixform} with respect to $\kernelcoefvec$
and $\samplealphavec$~\cite{micchelli2005function}
\begin{equation}
	(\kernelcoefvec, \samplealphaestvec) := \argmin_{\kernelcoefvec, \samplealphavec} \frac{1}{S} 
	\mathcal{L} (\boldmatvec v, \boldmatvec  y , 
	\samplekernelmat(\kernelcoefvec) \samplealphavec)
	+ \regpar \Omega((\samplealphavec^T \samplekernelmat(\kernelcoefvec) \samplealphavec)^{1/2})
	\label{eq:solutiongraphsmatrixformmkl}
\end{equation}
where $\samplekernelmat(\kernelcoefvec) \define \samplingmat
\fullkernelmat(\kernelcoefvec) \samplingmat^T$. Except for degenerate
cases, problem \eqref{eq:solutiongraphsmatrixformmkl} is not jointly
convex in $\kernelcoefvec$ and $ \samplealphaestvec$, but it is
separately convex in each vector for a convex
$\mathcal{L}$~\cite{micchelli2005function}. 
Iterative algorithms for solving~\eqref{eq:multi-1norm} and~\eqref{eq:solutiongraphsmatrixformmkl} are available in~\cite{romero2017kernel}.

\subsection{Semi-parametric reconstruction}
\cmt{3 pages}
\label{sec:semipar}
The approaches discussed so far are applicable to various problems but they are certainly limited 
by the modeling assumptions they make. 
In particular, the performance of algorithms belonging to the parametric 
family~\cite{narang2013localized,anis2016proxies,segarra2015percolation}
is restricted by how 
well the signals actually 
adhere to the selected model.
Nonparametric models on the other 
hand~\cite{kondor2002diffusion,smola2003kernels,chapelle2006,zamzam2016coupled}, offer 
flexibility and 
robustness but they cannot readily incorporate
information available a priori.   	

In practice however, it is not uncommon 
that neither of these 
approaches alone suffices for reliable inference. Consider, for instance, an employment-oriented social network such as LinkedIn, and suppose the goal is to 
estimate the salaries of all users given information about the salaries of a few.  Clearly, besides 
network connections, 
exploiting available information regarding the users' education level and work experience could 
benefit 
the 
reconstruction task. The same is true in problems arising in link analysis, where the exploitation of Web's hierarchical structure can aid the task of estimating the importance of Web pages~\cite{nikolakopoulos2013ncdawarerank}. In recommender 
systems, inferring preference scores for every item, given the users' feedback about particular 
items, could be cast as a 
signal reconstruction problem over the item correlation graph. Data sparsity 
imposes severe limitations in the quality of pure collaborative filtering 
methods~\cite{nikolakopoulos2017factored}
Exploiting side information about the items,   
is known to alleviate such limitations~\cite{nikolakopoulos2015top}, leading to considerably 
improved recommendation performance~\cite{nikolakopoulos2015hierarchical,6927541}.

A promising direction to endow nonparametric methods with prior information 
relies on a  \emph{semi-parametric} approach whereby the signal of interest 
is modeled as the superposition of 
a parametric and a nonparametric component~\cite{ioannidis2016semipar}. While the
former leverages side information, the latter accounts for deviations
from the parametric part, and can also promote smoothness using 
kernels on graphs. In this section we outline two simple and reliable semi-parametric estimators with complementary 
strengths, as detailed in \cite{ioannidis2016semipar}.

\subsubsection{Semi-parametric Inference}

Function $\signalfun$ is modeled as 
the superposition\footnote{for simplicity here we consider only the case of semi-parametric partially linear models.} 
\begin{align}
	\signalvec=\fpvec+\fnpvec
	\label{eq:decomp}
\end{align}
where $\fpvec:=[\fp(v_1),\ldots,\fp(v_\vertexnum)]\transpose$, and 
$\fnpvec:=[\fnp(v_1),$ $\ldots,\fnp(v_\vertexnum)]\transpose$.

The
parametric  term
$\fp(v):=\sum_{\paramind=1}^{{\paramnum}}\beta_{\paramind}\basisfun_{\paramind}(v)$
captures the known signal structure via the basis
$\mathcal{B}:=\{\basisfun_{\paramind}\}_{{\paramind}=1}^{\paramnum}$, while the nonparametric 
term
$\fnp$ belongs to an RKHS $\rkhs$, which accounts 
for deviations from the span of $\mathcal{B}$. 
The goal of this section is efficient and reliable 
estimation of $\signalvec$ given $\observationvec$, 
$\samplemat$, $\mathcal{B}$, $\rkhs$ and $\graph$.

Since $\fnp\in\rkhs$, vector $\fnpvec$ can be 
represented as in~\eqref{eq:generalform}. By defining 
$\paramcoefvec:=[\paramcoef_{1},\ldots,\paramcoef_{{\paramnum}}]\transpose$, and the $N
\times {\paramnum}$ matrix $\basismat$ with entries
$(\basismat)_{\vertexind,\paramind}:=\basisfun_{\paramind}(v_\vertexind)$, the parametric term can 
be written in vector form as $\fpvec\define\basismat\paramcoefvec$.
The semi-parametric estimates can be found as the solution of  
the 
following optimization problem 
\begin{align}
	\{\fullalphaestvec,\paramcoefestvec\} =\arg\min_{\fullalphavec,\paramcoefvec} 
	~~\frac{1}{\samplenum}&\sum_{\sampleind=1}^{\samplenum}
	\mathcal{L}(\observationfun_\sampleind,\signalfun(v_{\vertexind_\sampleind}))+ 
	\regpar 
	\|\fnp\|_\mathcal{H}^2\label{eq:reconstructionEq}\\
	\text{s.t.}~~~~~~\signalvec~&=\basismat\paramcoefvec+\fullkernelmat\fullalphavec\nonumber
\end{align} 
where the fitting loss $\mathcal{L}$ quantifies the deviation of $f$
from the data, and $\regpar >0$ is the regularization scalar that
controls overfitting the nonparametric term. 
Using 
(\ref{eq:reconstructionEq}), the semi-parametric estimates are 
expressed as 
$\signalestvec=\basismat\paramcoefestvec+\fullkernelmat\fullalphaestvec$.

Solving~\eqref{eq:reconstructionEq} entails minimization over 
$\vertexnum+\paramnum$ variables. Clearly, when dealing with large-scale graphs 
this could lead to prohibitively large computational cost. To reduce complexity, 
the semi-parametric version of the representer
theorem~\cite{scholkopf2002,scholkopf2001representer} is 
employed, which establishes 
that 
\begin{align}
	\signalestvec=\basismat\paramcoefestvec+
	\fullkernelmat\samplemat\transpose\samplealphaestvec
	\label{eq:vectorModel}
\end{align} 
where $\samplealphaestvec:=[\samplealphaest_{1},\ldots, 
\samplealphaest_{S}]\transpose$. Estimates $\samplealphaestvec,\paramcoefestvec$ 
are found as 
\begin{align}
	\{\samplealphaestvec,\paramcoefestvec\}=
	\arg\min_{\samplealphavec,\paramcoefvec} 
	~~\frac{1}{\samplenum}&\sum_{\sampleind=1}^{\samplenum}
	\mathcal{L} 
	(\observationfun_\sampleind,\signalfun(v_{\vertexind_\sampleind}))	 
	+\regpar 
	\|\fnp\|_\mathcal{H}^2\label{eq:samplereconstructionEq}\\
	\text{s.t.}~~~~~~\signalvec~&=\basismat\paramcoefvec+ 
	\fullkernelmat\samplemat\transpose\samplealphavec\nonumber
\end{align}
where 
$\samplealphavec:=[\samplealpha_{1},\ldots, 
\samplealpha_{S}]\transpose$. The RKHS norm 
in~\eqref{eq:samplereconstructionEq} is expressed as
$\|\fnp\|_\mathcal{H}^2= 
\samplealphavec\transpose\samplekernelmat
\samplealphavec$, with 
$\samplekernelmat\define\samplemat\fullkernelmat\samplemat\transpose$. 
Relative to (\ref{eq:reconstructionEq}), the number of optimization 
variables in (\ref{eq:samplereconstructionEq}) is reduced to the more affordable $\samplenum+{\paramnum}$, with  
$\samplenum\ll\vertexnum$. 

Next, two
loss functions with complementary benefits will be considered:
the \emph{square} loss and the 	\emph{$\epsilon$-insensitive} loss. The square loss function is 
\begin{align}
	\mathcal{L}(y_s,f(v_{n_s})):= \|y_s-f(v_{n_s}))\|_2^2
	\label{eq: least squares}
\end{align}
and (\ref{eq:samplereconstructionEq})  
admits 
the
following closed-form solution
\begin{subequations}
	\label{eq: leastsqsol}
	\begin{alignat}{2}
		\samplealphaestvec&=
		(\projectmat
		\samplekernelmat+\regpar\identitymat_\samplenum)\inv\projectmat 
		\observationvec
		\label{eq:alphasol}\\
		\label{eq:betasol}
		\paramcoefestvec&=(\samplebasismat\transpose\samplebasismat)\inv\samplebasismat
		\transpose(\observationvec-\samplekernelmat\samplealphaestvec)
	\end{alignat}
\end{subequations}
where $\samplebasismat\define\samplemat\basismat$,  $\projectmat\define 
\identitymat_\samplenum-\samplebasismat(\samplebasismat\transpose\samplebasismat
)\inv\samplebasismat\transpose$. 
The complexity of~\eqref{eq: leastsqsol} is 
$\mathcal{O}(\samplenum^3+\paramnum^3)$.

The 
\emph{$\epsilon$-insensitive} 
loss function 
is 	given by 
\begin{align}
	\mathcal{L}(y_s,f(v_{n_s})):=\max(0,|y_s-f(v_{n_s})|-\epsilon)
	\label{eq: huber loss}
\end{align}
where
$\epsilon$ is tuned, e.g. via cross-validation, to minimize the
generalization error and has well-documented merits 
in signal
estimation from quantized data~\cite{vapnik2013nature}.
Substituting~\eqref{eq: huber loss} into 
\eqref{eq:samplereconstructionEq} yields a convex non-smooth quadratic 
problem 
that can be solved efficiently  for 
$\samplealphavec$ and $\paramcoefvec $
using e.g. interior-point methods~\cite{scholkopf2002}.

\subsection{Numerical tests}
This section reports on the signal reconstruction performance of different methods  using real as well as  synthetic data. The performance 
of the estimators is assessed  
via Monte Carlo
simulation by comparing the normalized mean-square 
error (NMSE)
\begin{align}
	\text{NMSE}=\mathbb{E}\bigg{[}\frac{\|\signalestvec-\signalvec\|^2}{\|\signalvec\|^2}\bigg{]}.
\end{align}

\noindent {\bf Multi-kernel reconstruction.} The first data set contains 
departure 
and 
arrival 
information for
flights among U.S. airports~\cite{AirportsDataset}, from which 
$3\times 10^6$ flights in the months of July, August, and September of
2014 and 2015 were selected.  We construct a graph with  $\vertexnum = 50$ vertices corresponding to the airports with highest traffic, and whenever the number of flights between the two airports exceeds 100 within the observation
window, we connect the corresponding nodes with an edge.

A signal was constructed per day averaging the arrival delay
of \emph{all} inbound flights per selected airport. A total of
184 signals were considered, of which the first 154 were used for training
(July, August, September 2014, and July, August 2015), and the
remaining 30 for testing (September 2015).
The weights of the edges between airports were learned using the training data
based on the technique described in~\cite{romero2017kernel}.

\begin{table}[h!]
\centering
	  \title{Multi-Kernel Reconstruction}
\begin{tabular}{lcc}
\bottomrule
                     & NMSE  & RMSE[min] \\
\bottomrule
KRR with cov. Kernel & 0.34  & 3.95      \\
Multi-kernel, RS     & 0.44  & 4.51      \\
Multi-kernel, KS     & 0.43  & 4.45      \\
BL for B=2           & 1.55  & 8.45      \\
BL for B=3           & 32.64 & 38.72     \\
BL, cut-off          & 3.97  & 13.5      \\
\bottomrule
\end{tabular}
\caption{Multi-Kernel Reconstruction}
	\label{tab:airports}
\end{table}

Table~\ref{tab:airports} lists the NMSE and the RMSE in
minutes for the task of predicting the arrival delay at 40 airports
when the delay at a randomly selected collection of 10 airports is
observed.  The second row corresponds to the ridge regression
estimator that uses the nearly-optimal \emph{estimated} covariance
kernel. The next two rows correspond to the multi-kernel approaches
in \S\ref{sec:select_kernels} with a dictionary of 30 diffusion kernels 
with
values of $\sigma^2$ uniformly spaced between 0.1 and 7. The rest of the 
rows pertain to graph-bandlimited estimators (BL).
Table~\ref{tab:airports} demonstrates the reliable performance of
covariance kernels as well as the herein discussed multi-kernel approaches
relative to competing alternatives.

\setlength{\mywidth}{.35\textwidth}
 \setlength{\myheight}{.28\textwidth}
  \begin{figure}
  	\centering
  	\title{NMSE of the synthetic signal estimates.}
  	\subfigure[
  		($\regpar=5\times10^{-4}$, $\sigma=5\times10^{-4}$, 
  		$\text{SNR}_\observationnoisefun=5$dB).]{\label{fig:synthoutl}
  		\includegraphics[width=.47\linewidth]{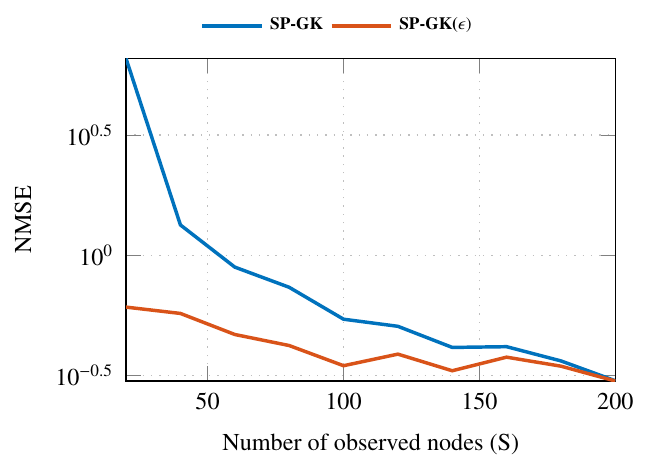}
  		}
  	~
  	\subfigure[
  		($\regpar=5\times10^{-4}$, $\sigma=5\times10^{-4}$, $\epsilon=10^{-4}$, and 
  		$\text{SNR}_{\observationoutlyingnoisefun}=-5$dB).]
  		{\label{fig:synth}
  		\includegraphics[width=.47\linewidth]{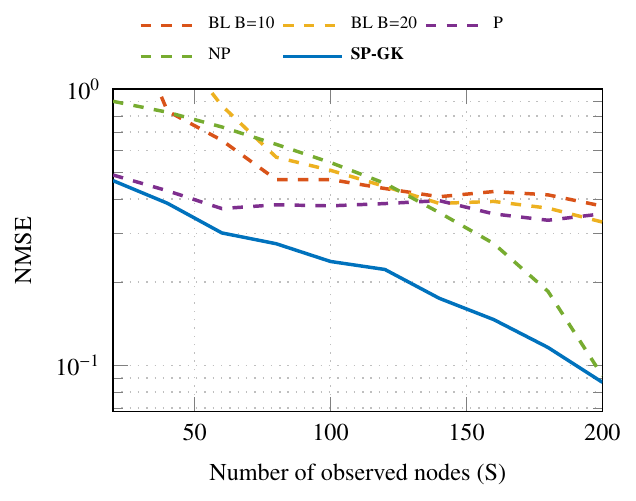}
  		}

  	\caption[Caption for LOF]{ NMSE of the synthetic signal estimates.
   		}
  \end{figure}
  
\newpage
  \noindent {\bf Semi-parametric reconstruction.} An 
  Erd\H{o}s-R$\grave{\text{e}}$nyi graph with probability of edge presence $0.6$
  and $\vertexnum=200$ nodes was generated, 
  and $\signalvec$
  was formed by superimposing a
  bandlimited signal~\cite{tsitsvero2016uncertainty,anis2016proxies} plus a
  piecewise constant signal~\cite{chen2015signal}; that is,
  \begin{align}
  	\signalvec = \sum_{i=1}^{10}\gamma_i \boldmatvec u_i +
  	\sum_{i=1}^{6}\delta_i\boldmatvec{1}_{\localset_c}
  \end{align}where
  $\{\gamma_i\}_{i=1}^{10}$ and $\{\delta_i\}_{i=1}^{6}$ are standardized
  Gaussian distributed; $\{\boldmatvec u_i\}_{i=1}^{10}$ are the eigenvectors
  associated with the 10 smallest eigenvalues of the Laplacian matrix;
  $\{\localset_i\}_{i=1}^6$ are the vertex sets of $6$ clusters obtained
  via spectral clustering~\cite{von2007tutorial}; and $\boldmatvec 1_{\localset_i}$ is the
  indicator vector with entries ${{(\boldmatvec{1}_{\localset_i})}_n}:=1$, if
  $v_n \in \localset_i$, and $0$ otherwise. 
  The parametric basis 
  $\mathcal{B}=\{\boldmatvec{1}_{\localset_i}\}_{i=1}^6$ was used by the estimators  capturing 
  the prior knowledge, and $S$
  vertices were sampled uniformly at random. 
  The subsequent experiments evaluate the performance of the
  semi-parametric 
  graph kernel estimators, SP-GK and 
  SP-GK($\epsilon$) resulting from using~\eqref{eq: least 
  	squares} 
  and~\eqref{eq: huber loss} in~\eqref{eq:samplereconstructionEq}, 
  respectively; the \emph{parametric} (P) 
  that considers only the parametric term in~\eqref{eq:decomp}; 
  the \emph{nonparametric}
  (NP)~\cite{kondor2002diffusion,smola2003kernels} that 
  considers 
  only the 
  nonparametric term 
  in~\eqref{eq:decomp}; 
  and  the graph-bandlimited estimators (BL)
  from~\cite{tsitsvero2016uncertainty,anis2016proxies}, which
  assume a bandlimited model with bandwidth  $\bandwidth$. For 
  all 
  the 
  experiments, 
  the diffusion 
  kernel (cf. Table~\ref{tab:spectralweightfuns}) with 
  parameter~$\sigma$ is employed. First, white Gaussian noise 
  $\observationnoisefun_\sampleind$ 
  of variance
  $\sigma_\observationnoisefun^2$ is added to each sample $\signalfun_\sampleind$ to yield 
  signal-to-noise ratio 
  $\text{SNR}_\observationnoisefun:={\|\signalvec\|_2^2}/({N\sigma_\observationnoisefun^2}) 
  $.   
  Fig.~\ref{fig:synth} presents the
  NMSE of different methods. 
  As expected, the limited flexibility 
  of the parametric approaches, BL
  and P, affects their ability to capture the 
  true 
  signal structure. The NP estimator achieves 
  smaller NMSE, but only when the 
  amount of available samples is adequate. 
  Both semi-parametric estimators were found to 
  outperform other approaches, exhibiting reliable 
  reconstruction even with few samples.

  To illustrate the benefits of employing different loss 
  functions~\eqref{eq: least 
  	squares} 
  and~\eqref{eq: huber loss}, we compare the 
  performance of 
  SP-GK 
  and 
  SP-GK($\epsilon$) in the presence of outlying noise.  Each sample $\signalfun_\sampleind$ is 
  contaminated with Gaussian noise $\observationoutlyingnoisefun_\sampleind$ of large 
  variance 
  $\sigma^2_{\observationoutlyingnoisefun}$ with probability 
  $p=0.1$.  Fig.~\ref{fig:synthoutl} demonstrates the robustness of SP-GK($\epsilon$)  
  which is attributed to the $\epsilon-$insensitive loss 
  function~\eqref{eq: huber loss}.
  Further experiments using real signals can be found in 
  \cite{ioannidis2016semipar}.
  
\section{Inference of dynamic functions over dynamic graphs}
\label{sec:timevar}
Networks that exhibit \textit{time-varying connectivity} patterns with \textit{time-varying node attributes} 
arise in a plethora of network science related applications. 
Sometimes these dynamic network topologies switch between a finite
number of discrete states, governed 
by sudden changes of the underlying 
dynamics~\cite{shen2016nonlinear,baingana2017tracking}.
A challenging problem that arises in this setting is that of 
	reconstructing time-evolving functions on graphs, 
	given 
	their values on a subset
	of vertices and time instants. Efficiently
	exploiting spatiotemporal dynamics can markedly impact sampling costs
	by reducing the number of vertices that need to be observed to attain
	a target performance. Such a reduction can be of
	paramount importance in certain applications eg. in monitoring 
	time-dependent activity of different regions 
	of the brain through 
	invasive
	electrocorticography (ECoG), where observing a vertex requires the
	implantation of an intracranial electrode~\cite{shen2016nonlinear}. 	

Although one could
reconstruct a time-varying function per time slot using the non- or semi-parametric methods of~\S\ref{sec:tirecon},  
leveraging time 
correlations typically 
yields estimators with improved performance.  Schemes tailored for
	time-evolving  functions on graphs 
include~\cite{bach2004learning} and \cite{mei2016causal},
		which 
		 predict the function values at time
			$\timeind$ given observations up to time $\timeind-1$.
			However, these schemes assume that
			the function of interest adheres to a specific vector
			autoregressive model. 
 Other works target
	time-invariant functions, but can only afford tracking sufficiently slow
	variations. This is the case with the dictionary learning approach
	in~\cite{forero2014dictionary} and the distributed algorithms
	in~\cite{wang2015distributed} and
	\cite{lorenzo2016lms}. Unfortunately, the flexibility of these
	algorithms to capture spatial information is also limited
	since~\cite{forero2014dictionary} focuses on Laplacian
	regularization, whereas~\cite{wang2015distributed} and
	\cite{lorenzo2016lms} require the signal to be bandlimited. 
	
Motivated by the aforementioned limitations, in what comes next we extend the framework presented in \S\ref{sec:tirecon} accommodating time-varying function reconstruction over dynamic graphs. But before we delve into the time-varying setting, a few definitions are in order. 

 \noindent{\textbf{Definitions}:} A time-varying
	graph 
	is a tuple
	$\graph\timescalarnot{\timeind}:=(\vertexset,\spaceadjacencymat
	\timenot{\timeind})$,
	where $\vertexset:=\{v_1, \ldots, v_\vertexnum\}$ is the vertex set, 
	and $\spaceadjacencymat\timenot{\timeind}\in
	\rfield^{\vertexnum\times\vertexnum}$ is the adjacency matrix at
	time $\timeind$, whose $(\vertexind,\vertexindp)$-th entry
	$\spaceadjacencymatentry\timevertexvertexnot{\timeind}{\vertexind}{\vertexindp}$
	assigns a weight to the pair of vertices
	$(v_\vertexind,v_\vertexindp)$ at time $\timeind$. A time-invariant
	graph is a special case with $\spaceadjacencymat\timenot{\timeind}=
	\spaceadjacencymat\timenot{\timeindp}~\forall\timeind,\timeindp$.
	{Adopting common assumptions made in related literature (e.g.~\cite[Ch. 
		2]{kolaczyck2009},\cite{shuman2013emerging,belkin2006manifold}}),
	 we also define $\graph\timescalarnot{\timeind}$ (i) to have  non-negative 
	weights
	($\spaceadjacencymatentry\timevertexvertexnot{\timeind}{\vertexind}{\vertexindp}\geq
	0~\forall \timeind, \text{ and } \forall \vertexind \neq \vertexindp$); (ii) to have no self-edges
	($\spaceadjacencymatentry\timevertexvertexnot{\timeind}{\vertexind}{\vertexind}=0~\forall\vertexind,\timeind$);
	and, (iii) to be undirected
	($\spaceadjacencymatentry\timevertexvertexnot{\timeind}{\vertexind}{\vertexindp}=
	\spaceadjacencymatentry\timevertexvertexnot{\timeind}{\vertexindp}{\vertexind}~\forall
	\vertexind,\vertexindp,\timeind$).

	A time-varying function or signal on a
	graph is a map $
	\signalfun:\vertexset\times\timeset \rightarrow \mathbb{R}$, where
	$\timeset\define\{1,2,\ldots\}$ is the set of time
	indices. The value $\signalfun(v_\vertexind,\timeind)$ of
	$\signalfun$ at vertex $v_\vertexind$ and time $\timeind$, 
	can be thought of as the value of an attribute of $v_\vertexind\in
	\vertexset$ at time $\timeind$. 
	The
	values of $\signalfun$ at time $\timeind$ will be collected in
	$\signalvec\timenot{\timeind}:=[\signalfun\timevertexnot{\timeind}{1},\ldots,
	\signalfun\timevertexnot{\timeind}{N}]\transpose$.

At time $\timeind$,  vertices with
	indices in the time-dependent set
	$\sampleset\timenot{\timeind}:=\{\vertexind\sampletimenot{1}{\timeind}, 
	\ldots,\vertexind\sampletimenot{\samplenum\timescalarnot{\timeind}}{\timeind}\}$,
	$1\leq
	\vertexind\sampletimenot{1}{\timeind}<\cdots<\vertexind\sampletimenot{\samplenum\timescalarnot{\timeind}}{\timeind}\leq
	\vertexnum$, are observed. The resulting samples can be expressed as
	$\observationfun\sampletimenot{\sampleind}{\timeind} =
	\signalfun\timevertexnot{\timeind}{\vertexind_\sampleind\timescalarnot{\timeind}}
	+ \observationnoisefun\sampletimenot{\sampleind}{\timeind},
	\sampleind=1,\ldots,\samplenum\timescalarnot{\timeind}$, where
	$\observationnoisefun\sampletimenot{\sampleind}{\timeind}$ models
	observation error. 
	%
	By letting
	$\observationvec\timenot{\timeind} \define
	[\observationfun\sampletimenot{1}{\timeind},\ldots,\observationfun\sampletimenot{\samplenum\timescalarnot{\timeind}}{\timeind}]\transpose$,
	the observations can be conveniently expressed as
	\begin{equation}
		\label{eq:observationsvectv}
		\observationvec\timenot{\timeind} =\samplemat\timenot{\timeind} 
		\signalvec\timenot{\timeind}  + 
		\observationnoisevec\timenot{\timeind},\quad\timeind=1,2,\ldots
	\end{equation}
	where $\observationnoisevec\timenot{\timeind} \define
	[\observationnoisefun\sampletimenot{1}{\timeind},\ldots,\observationnoisefun\sampletimenot{\samplenum\timescalarnot{\timeind}}{\timeind}]\transpose$,
	and the $\samplenum\timescalarnot{\timeind}\times \vertexnum$ sampling
	matrix $\samplemat\timenot{\timeind}$ contains ones at positions
	$(\sampleind,\vertexind\sampletimenot{\sampleind}{\timeind})$,
	$\sampleind=1,\ldots,\samplenum\timescalarnot{\timeind}$ and zeros
	elsewhere. 
	
	The broad goal of this section is to
	``reconstruct'' $\signalfun$ from the observations
	$\{\observationvec\timenot{\timeind}\}_{\timeind}$ in
	\eqref{eq:observationsvectv}. Two formulations will be considered.
	
	\noindent {\bf Batch formulation.} In 
	the batch 
	reconstruction
	problem, one aims at 	 finding $\{\signalvec\timenot{\timeind}\}_{\timeind=1}^\timenum$
		 given $\{\graph\timescalarnot{\timeind}\}_{\timeind=1}^\timenum$, the sample locations
		$\{\samplemat\timenot{\timeind}\}_{\timeind=1}^\timenum$, and all
		observations
		$\{\observationvec\timenot{\timeind}\}_{\timeind=1}^\timenum$.

	\noindent {\bf Online formulation.} 
	At every time $\timeind$, one is
	given $\graph$ together with $\samplemat\timenot{\timeind}$ and
	$\observationvec\timenot{\timeind}$, and the goal is to find
	$\signalvec\timenot{\timeind}$. The latter can be obtained possibly
	based on a previous estimate of $\signalvec\timenot{\timeind-1}$, but
	the complexity per time slot $\timeind$ must be independent of $t$. 

\cmt{Prior info=smoothness}To solve these problems, 
we will rely on the
assumption that $\signalfun$ evolves smoothly over space and
time, yet more structured dynamics can be incorporated if known.

\subsection{Kernels on extended graphs} 
\label{sec:reconstruction}
This section extends the kernel-based
learning framework of \S\ref{sec:tirecon} to subsume time-evolving
functions over possibly dynamic graphs through the notion of
\emph{graph extension}, by which the time dimension receives the
same treatment as the spatial dimension.  The versatility of
kernel-based methods to leverage spatial
information~\cite{romero2017kernel} is thereby inherited and
broadened to account for temporal dynamics as well.  
This vantage point also accommodates time-varying sampling sets and
topologies. 

\subsubsection{Extended graphs} 
An immediate approach to reconstructing
time-evolving functions is to apply \eqref{eq:solutiongraphsf} separately
for each $\timeind=1,\ldots,\timenum$. This yields the instantaneous
estimator (IE)
\begin{align}
	\label{eq:timeagnostic}
	\signalestvec^{\text({IE})}\timenot{\timeind}
	:=& \argmin_{\rkhsvec} \frac{1}{\samplenum\timescalarnot{\timeind}} 
	||\observationvec\timenot{\timeind} - \samplemat\timenot{\timeind}\rkhsvec||^2_2
	+ \regpar \rkhsvec\transpose \fullkernelmat\pinv\timenot{\timeind}
	\rkhsvec.
\end{align}
\cmt{critic=no time info}Unfortunately, this estimator does not
account for the possible relation between e.g.
$\signalfun\timevertexnot{\timeind}{\vertexind}$ and
$\signalfun\timevertexnot{\timeind-1}{\vertexind}$. If, for instance,
$\signalfun$ varies slowly over time, an estimate of
$\signalfun\timevertexnot{\timeind}{\vertexind}$ may as well benefit
from leveraging observations
$\observationfun\sampletimenot{\sampleind}{\timeindaux}$ at time
instants $\timeindaux\neq\timeind$. Exploiting temporal dynamics
potentially reduces the number of vertices that have to be sampled to
attain a target reconstruction performance, which in turn can markedly
reduce sampling costs.

\cmt{Graph extension}
Incorporating temporal dynamics into
	kernel-based reconstruction, which can only handle a single snapshot
	(cf.  \S\ref{sec:tirecon}), necessitates an appropriate
	reformulation of time-evolving function reconstruction as a problem
	of reconstructing a time-invariant function. An appealing
	possibility is to replace $\graph$ with its \emph{extended} version  
	$\extendedgraph\define( \extendedvertexset,\extendedadjacencymat)$,
	where each vertex in $\vertexset$ is replicated $\timenum$ times to
	yield the extended vertex set
	$\extendedvertexset\define\{v\vertextimenot{\timeind}{\vertexind},~\vertexind=1,\ldots,\vertexnum,~\timeind=1,\ldots,\timenum\}$,
	and the
	$(\vertexind+\vertexnum(\timeind-1),\vertexindp+\vertexnum(\timeindp-1))$-th
	entry of the $\timenum\vertexnum\times\timenum\vertexnum$ extended
	adjacency matrix $\extendedadjacencymat$ equals the weight of the
	edge
	$(v\vertextimenot{\timeind}{\vertexind},v\vertextimenot{\timeindp}{\vertexindp})$.
	The
	time-varying function $\signalfun$ can thus be replaced with its extended
	time-invariant counterpart
	$\extendedsignalfun:\extendedvertexset\rightarrow\rfield$ 
	with  
	$\extendedsignalfun(v\vertextimenot{\timeind}{\vertexind})=\signalfun\timevertexnot{\timeind}{\vertexind}$.

			\begin{mydefinitionhere}
				Let   $\vertexset:=\{v_1, \ldots,
				v_\vertexnum\}$ denote a vertex set and let
				$\graph:=(\vertexset,\{\spaceadjacencymat\timenot{\timeind}\}_{\timeind=1}^{\timenum})$
				be a time-varying graph. A graph $\extendedgraph$ with vertex 
				set
				$\extendedvertexset\define\{v\vertextimenot{\timeind}{\vertexind},~\vertexind=1,\ldots,\vertexnum,~\timeind=1,\ldots,\timenum\}$
				and $\vertexnum\timenum\times\vertexnum\timenum$ adjacency 
				matrix
				$\extendedadjacencymat$ is an extended graph of $\graph$ if the
				$\timeind$-th $\vertexnum\times\vertexnum$ diagonal block of
				$\extendedadjacencymat$ equals
				$\spaceadjacencymat\timenot{\timeind}$.
			\end{mydefinitionhere}
			In general, the diagonal blocks $\{\spaceadjacencymat\timenot{\timeind}\}_{\timeind=1}^{\timenum}$ do not provide full description of the underlying extended graph. Indeed, one also needs to specify the off-diagonal block entries of $\extendedadjacencymat$ to capture the spatio-temporal dynamics of 	$\signalfun$. 

As an example,
		consider an extended graph with
		\begin{align}
			\label{eq:timevaryingextendedadjacencymat}
			&  \extendedadjacencymat=
			\tridiag\{\spaceadjacencymat\timenot{1},\ldots,\spaceadjacencymat\timenot{\timenum};\timeconnectionmat\timenot{2},\ldots,\timeconnectionmat\timenot{\timenum}\}
		\end{align}
		where
		$\timeconnectionmat\timenot{\timeind}\in\rfield_+^{\vertexnum\times\vertexnum}$
		connects 
		$\{v\vertextimenot{\timeind-1}{\vertexind}\}_{\vertexind=1}^\vertexnum$
		to
		$\{v\vertextimenot{\timeind}{\vertexind}\}_{\vertexind=1}^\vertexnum$,
		$\timeind=2,\ldots,\timenum$ and 
		$ \tridiag\{\boldmatvec A_1,$  $\ldots,\boldmatvec 
		A_\timenum;\boldmatvec B_2,\ldots,\boldmatvec B_\timenum\}$
represents the symmetric block tridiagonal matrix:
\begin{displaymath}
\extendedadjacencymat=\left[ 
    \begin{array}{cccccc}
      \boldmatvec A_{1}  &
      \boldmatvec B\transpose _{2} &\boldmatvec0 &\ldots & \boldmatvec0 & 
      \boldmatvec0 \\
      \boldmatvec B_{2} & \boldmatvec A_{2} &\boldmatvec B\transpose_{3} & 
      \ldots &\boldmatvec0 & \boldmatvec0 \\
      \boldmatvec0 & \boldmatvec B_{3} & \boldmatvec A_{3} &\ldots & 
      \boldmatvec0 & \boldmatvec 0\\
      \vdots & \vdots & \vdots & \ddots & \vdots & \vdots\\
      \boldmatvec0 & \boldmatvec0 & \boldmatvec0 &\ldots &\boldmatvec 
      A_{\timenum-1}& \boldmatvec B\transpose_{\timenum}\\
      \boldmatvec0 & \boldmatvec0 & \boldmatvec0 &\ldots & \boldmatvec 
      B_{\timenum} &\boldmatvec A_{\timenum}\\
    \end{array}
  \right].
  \end{displaymath}

For instance, each vertex can be connected 
		to its neighbors at the previous time instant by setting
		$\timeconnectionmat\timenot{\timeind}=\spaceadjacencymat\timenot{\timeind-1}$,
		or it can be connected to its replicas at adjacent time
		instants by setting $\timeconnectionmat\timenot{\timeind}$ to be
		diagonal. 

	\begin{figure}
 		\centering
		\includegraphics[width=.50\textwidth]{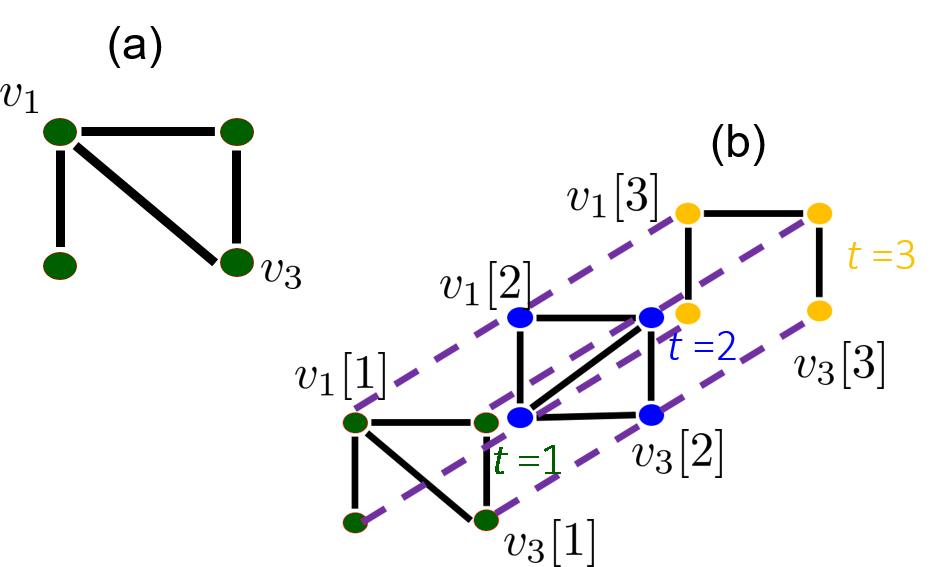}
		\caption{(a)Original graph (b) Extended graph
			$\extendedgraph$ for
			diagonal~$\timeconnectionmat\timenot{\timeind}$. {
			Edges connecting
				vertices at the same time instant are represented by solid lines
				whereas edges connecting vertices at different time instants are
				represented by dashed lines. 
					}
		}\label{fig:extendedgraph}
	\end{figure}

\subsubsection{Batch and online reconstruction via space-time kernels}
\cmt{Estimation}The extended graph 
enables a generalization of
the estimators in \S\ref{sec:tirecon} for 
time-evolving functions. The rest of this subsection discusses two such KRR
estimators.

Consider first the batch
		formulation, where all the
	$\extendedsamplenum\define\sum_{\timeind=1}^\timenum\samplenum
	\timenot{\timeind}$ samples in $\extendedobservationvec\define [
	\observationvec\transpose\timenot{1},\ldots,\observationvec\transpose\timenot{\timenum}
	]\transpose$ are available, and the goal is to estimate
	$\extendedrkhsvec\define[\rkhsvec\transpose\timenot{1},\ldots,\rkhsvec\transpose\timenot{\timenum}]\transpose$.
	Directly
	applying the KRR criterion in~\eqref{eq:solutiongraphsf} to reconstruct
	$\extendedsignalvec$ on the extended graph $\extendedgraph$ yields
	\begin{subequations}
		\begin{equation}
			\label{eq:batchcriterion}
			\extendedsignalestvec
			:= \argmin_{\extendedrkhsvec} 
			||\extendedobservationvec - 
			\extendedsamplemat\extendedrkhsvec||^2_{\samplenumdiagmat}
			+ \regpar \extendedrkhsvec\transpose \extendedfullkernelmat\pinv 
			\extendedrkhsvec
		\end{equation}
		where
		 $ \extendedfullkernelmat$ is now a
			$\timenum\vertexnum\times\timenum\vertexnum$ ``space-time'' kernel
			matrix, 
			
			$\extendedsamplemat\define\bdiag{\samplemat\timenot{1},\ldots,\samplemat\timenot{\timenum}
			}$, and
			 $\samplenumdiagmat \define \bdiag{
				\samplenum\timescalarnot{1}\boldmatvec 
				I_{\samplenum\timescalarnot{1}}
				,
				\ldots,
				\samplenum\timescalarnot{\timenum}\boldmatvec 
				I_{\samplenum\timescalarnot{\timenum}}
			}$. 
		If 
		$\extendedfullkernelmat$ is invertible, \eqref{eq:batchcriterion} can 
		be solved in
		closed form as
		\begin{align}
			\extendedsignalestvec=
			\extendedfullkernelmat
			\extendedsamplemat\transpose
			(
			\extendedsamplemat
			\extendedfullkernelmat
			\extendedsamplemat\transpose
			+
			\regpar \samplenumdiagmat
			)\inv
			\extendedobservationvec.
			\label{eq:closedbatchform}
		\end{align}
	\end{subequations} 
	The ``space-time" kernel $\extendedfullkernelmat$,  
		captures complex spatiotemporal dynamics. If the topology is time 
		invariant, 
		$\extendedfullkernelmat$ can be specified in a bidimensional plane of 
		spatio-temporal
		frequency similar to \S\ref{sec:kernelsong}\footnote{For general designs of 
		space-time kernels
		$\extendedfullkernelmat$ for time-invariant as well as time-varying 
		topologies see~\cite{romero2016spacetimekernel}.}.

\label{sec:onlineestimator}

In the online formulation,
	one aims to estimate 
	$\signalvec\timenot{\timeind}$ after the
	$\extendedsamplenum\timescalarnot{\timeind}\define\sum_{\timeindaux=1}^\timeind\samplenum
	\timescalarnot{\timeindaux}$ samples in
	$\extendedobservationvec\timenot{\timeind}\define [
	\observationvec\transpose\timenot{1},\ldots,\observationvec\transpose\timenot{\timeind}
	]\transpose$ become available. Based on these samples, the KRR estimate of
	$\extendedsignalvec$, denoted as 
	$\extendedsignalestvec\timeperiodgiventimenot{\timeind}$,
	is 
	clearly
	\begin{subequations}
		\label{eq:onlinecriteriongeneral}
		\begin{eqnarray}
			\extendedsignalestvec\timeperiodgiventimenot{\timeind}
			:=& \argmin_{\extendedrkhsvec} 
			||\extendedobservationvec\timenot{\timeind} -
			\extendedsamplemat\timenot{\timeind}\extendedrkhsvec||^2_{
				\samplenumdiagmat\timenot{\timeind}
			}
			+ \regpar \extendedrkhsvec\transpose
			\extendedfullkernelmat\inv
			\extendedrkhsvec
			\label{eq:onlinecriterion}\\
			=&
			\extendedfullkernelmat
			\extendedsamplemat\transpose\timenot{\timeind}
			(
			\extendedsamplemat\timenot{\timeind}
			\extendedfullkernelmat
			\extendedsamplemat\transpose\timenot{\timeind}
			+
			\regpar \samplenumdiagmat\timenot{\timeind}
			)\inv
			\extendedobservationvec\timenot{\timeind}.
			\label{eq:onlinecriterionsol}
		\end{eqnarray}
	\end{subequations}
	where $\extendedfullkernelmat$ is assumed invertible 
		for simplicity,
		$\samplenumdiagmat\timenot{\timeind}\define \bdiag{
			\samplenum\timescalarnot{1}\boldmatvec 
			I_{\samplenum\timescalarnot{1}} , \ldots,
			\samplenum\timescalarnot{\timeind}\boldmatvec 
			I_{\samplenum\timescalarnot{\timeind}} }$,
		and
		$\extendedsamplemat\timenot{\timeind}\define[\diag{\samplemat\timenot{1},\ldots,\samplemat\timenot{\timeind}},\boldmatvec0_{
			\extendedsamplenum\timescalarnot{\timeind} \times
			(\timenum-\timeind)\vertexnum }]
		\in\{0,1\}^{\extendedsamplenum\timescalarnot{\timeind}\times
			\timenum\vertexnum}$.
	
	 The estimate in \eqref{eq:onlinecriteriongeneral} comprises
		the per slot estimates
		$\{\signalestvec\timegiventimenot{\tau}{\timeind}\}_{\tau=1}^\timenum$;
		that is, $ \extendedsignalestvec\timeperiodgiventimenot{\timeind} 
		\define[
		\signalestvec\transpose\timegiventimenot{1}{\timeind},
		\ldots,$ $
		\signalestvec\transpose\timegiventimenot{\timenum}{\timeind}
		]\transpose $ with $
		\signalestvec\timegiventimenot{\tau}{\timeind}:=[
		\signalestfun\timegiventimevertexnot{\tau}{\timeind}{1}, \ldots,
		\signalestfun\timegiventimevertexnot{\tau}{\timeind}{\vertexnum}]\transpose$,
		where $\signalestvec\timegiventimenot{\tau}{\timeind}$ (respectively
		$ \signalestfun\timegiventimevertexnot{\tau}{\timeind}{\vertexind}$)
		is the KRR estimate of $\signalvec\timenot{\tau}$
		($\signalfun\timevertexnot{\tau}{\vertexind}$) given the
		observations up to time $\timeind$. With this notation, 
		it follows
		that for all $\timeind,\tau$
		\begin{align}
			\label{eq:onlinecriteriontimet}
			\signalestvec\timegiventimenot{\tau}{\timeind} &=
			(\canonicalvec{\timenum}{\tau}\transpose\otimes
			\identitymat_{\vertexnum})
			\extendedsignalestvec\timeperiodgiventimenot{\timeind}.
		\end{align}

	{Regarding $\timeind$ as the present,
		\eqref{eq:onlinecriteriongeneral} therefore provides estimates of
		past, present, and future values of $f$. The solution to the online
		problem 
		comprises the
		sequence of \emph{present} KRR estimates for all $t$, that is,
		$\{\signalestvec\timegiventimenot{\timeind}{\timeind}\}_{\timeind=1}^\timenum$.
		This can be obtained by solving \eqref{eq:onlinecriterion} in closed
		form per $\timeind$ as in \eqref{eq:onlinecriterionsol}, and then
		applying \eqref{eq:onlinecriteriontimet}. However, such an approach
		does not yield a desirable online algorithm since its complexity per
		time slot is 
		$\mathcal{O}(\extendedsamplenum^3\timescalarnot{\timeind})$, and 
		therefore increasing with $t$.
		For this reason, this approach is not satisfactory since the online
		problem formulation 
		requires the
		complexity per time slot of the desired algorithm to be independent of $\timeind$.  An
		algorithm that does satisfy this requirement 
		yet
		provides the exact KRR estimate is presented next when 
		the kernel matrix is any positive definite  matrix $
		\extendedfullkernelmat$ satisfying}
	\begin{align}
		\label{eq:invblocktriagonal}
		\extendedfullkernelmat\inv&=
		\tridiag\{\kernelinvondiagonalmat\timenot{1},\ldots,\kernelinvondiagonalmat\timenot{\timenum}
		;\kernelinvoffdiagonalmat\timenot{2},\ldots,\kernelinvoffdiagonalmat\timenot{\timenum}
		\}
	\end{align}
	for some $\vertexnum\times\vertexnum$ matrices
	$\{\kernelinvondiagonalmat\timenot{\timeind}\}_{\timeind=1}^\timenum$
	and
	$\{\kernelinvoffdiagonalmat\timenot{\timeind}\}_{\timeind=2}^\timenum$.
	Kernels in this important family are designed 
	in~\cite{romero2016spacetimekernel}.
	
	If 
	$\extendedfullkernelmat$ is of the form
	\eqref{eq:invblocktriagonal} 
	then the kernel Kalman filter (KKF) in
	Algorithm~\ref{algo:kalmanfilter} returns the sequence
	$\{\signalestvec\timegiventimenot{\timeind}{\timeind}\}_{\timeind=1}^\timenum$,
	where $\signalestvec\timegiventimenot{\timeind}{\timeind}$ is
	given by~\eqref{eq:onlinecriteriontimet}. The 
	$\vertexnum\times\vertexnum$ matrices 
	$\{\transitionmat\timenot{\tau}\}_{\tau=2}^\timenum$
	and~$\{\plantnoisekernelmat\timenot{\tau}\}_{\tau=1}^{\timenum}$
	are obtained offline by Algorithm~\ref{algo:rec}, and
	$\observationnoisevar\timescalarnot{\tau}= 
	\regpar\samplenum\timescalarnot{\tau}~\forall\tau$.
	
	{The KKF generalizes
		the probabilistic KF since the latter is recovered upon setting
		$\extendedfullkernelmat$ to be the covariance matrix of
		$\extendedrkhsvec$ in the probabilistic KF. 
		The assumptions made by
		the probabilistic KF are stronger than those involved in the
		KKF. Specifically, in the probabilistic KF,
		$\rkhsvec\timenot{\timeind}$ must adhere to a linear state-space
		model,  $\signalvec\timenot{\timeind}=\transitionmat\timenot{\timeind}
		\signalvec\timenot{\timeind-1} + \plantnoisevec\timenot{\timeind}$, with known transition 
		matrix
		$\transitionmat\timenot{\timeind}$, where the state noise
		$\plantnoisevec\timenot{\timeind}$ is uncorrelated over time and has
		known covariance matrix
		$\plantnoisekernelmat\timenot{\timeind}$. Furthermore, the
		observation noise $\observationnoisevec\timenot{\timeind}$ must be
		uncorrelated over time and have known covariance
		matrix. Correspondingly, the performance guarantees of the
		probabilistic KF are also stronger: the resulting estimate is
		optimal in the mean-square error sense among all linear
		estimators. Furthermore, if $\plantnoisevec\timenot{\timeind}$ and
		$\observationvec\timenot{\timeind}$ are jointly Gaussian,
		$\timeind=1,\ldots,\timenum$, then the probabilistic KF estimate is
		optimal in the mean-square error sense among all (not necessarily
		linear) estimators.}  {In contrast, the requirements of the proposed
		KKF are much weaker since they only specify that $\signalfun$ must
		evolve smoothly with respect to a given extended graph. As expected,
		the performance guarantees are similarly weaker; see
		e.g.~\cite[Ch. 5]{scholkopf2002}. However, since the KKF generalizes
		the probabilistic KF, the reconstruction performance of the former
		for judiciously selected $\extendedfullkernelmat$ cannot be worse
		than the reconstruction performance of the latter for any given
		criterion. The caveat, however, is that such a selection is not
		necessarily~easy. }
	
	For the rigorous statement and proof  relating the celebrated 
	KF~\cite[Ch. 
	17]{strang1997linear} and the optimization problem 
	in~\eqref{eq:onlinecriterion}, 
	see~\cite{romero2016spacetimekernel}. 
	Algorithm~\ref{algo:kalmanfilter}
	requires $\mathcal{O}(\vertexnum^3)$ operations per time slot, whereas
	the complexity of evaluating \eqref{eq:onlinecriterionsol} for the
	$\timeind$-th time slot is
	$\mathcal{O}(\extendedsamplenum^3\timescalarnot{\timeind})$, which increases
	with $\timeind$ and becomes eventually prohibitive. Since distributed 
	versions of the Kalman filter are well 
	studied~\cite{schizas2008consensus}, decentralized KKF algorithms can 
	be pursued  to further reduce the computational complexity.

	\begin{algorithm}[t]
		\caption{Recursion to set the parameters of the KKF}\label{algo:rec}
		\textbf{Input:} $\kernelinvondiagonalmat\timenot{\timeind}$, $\timeind =
		1,\ldots,\timenum$, 
		$\kernelinvoffdiagonalmat\timenot{\timeind}$, $\timeind =
		2,\ldots,\timenum$.\\
		\begin{algorithmic}[1]
			\STATE  \textbf{Set}  
			$\plantnoisekernelmat\inv\timenot{\timenum}=\kernelinvondiagonalmat\timenot{\timenum}$\\
			
			\STATE \textbf{for} {$\timeind = \timenum,~\timenum-1,\ldots,2$} 
			\textbf{do}
			\STATE\quad    
			$\transitionmat\timenot{\timeind}=-\plantnoisekernelmat\timenot{\timeind}\kernelinvoffdiagonalmat\timenot{\timeind}$\\
			\STATE\quad    
			$\plantnoisekernelmat\inv\timenot{\timeind-1}=\kernelinvondiagonalmat\timenot{\timeind-1}-\transitionmat\transpose\timenot{\timeind}
			\plantnoisekernelmat\inv\timenot{\timeind}\transitionmat\timenot{\timeind}$\\
			%
			%
			%
			%
		\end{algorithmic}
		\textbf{Output: } $\plantnoisekernelmat\timenot{\timeind}$, $\timeind =
		1,\ldots,\timenum$, 
		$\transitionmat\timenot{\timeind}$, $\timeind =
		2,\ldots,\timenum$\\
	\end{algorithm}

	\begin{algorithm}[t]                
		\caption{Kernel Kalman filter (KKF)}
		\label{algo:kalmanfilter}    
		\begin{minipage}{20cm}
			\textbf{Input:} 
			$\{\plantnoisekernelmat\timenot{\timeind}\in\pdset^{\vertexnum}\}_{\timeind
				=
				1}^\timenum$,  $\{\transitionmat\timenot{\timeind}\in 
			\rfield^{\vertexnum\times\vertexnum}\}_{\timeind =
				2}^\timenum$, \\
			\indent\hspace{1cm}$\{\observationvec\timenot{\timeind}\in\rfield^{\samplenum\timescalarnot{\timeind}}\}_{\timeind=1}^\timenum$,
			$\{\samplemat\timenot{\timeind}\in\{0,1\}^{\samplenum\timescalarnot{\timeind}\times\vertexnum}\}_{\timeind=1}^\timenum$,
			\\
			\indent\hspace{1cm}$\{\observationnoisevar\timescalarnot{\timeind}>0\}_{\timeind=1}^\timenum$.
			\begin{algorithmic}[1]
				\STATE\textbf{Set} $\signalestvec\timenot{0|0}=\boldmatvec 0$,
				$\errormat\timegiventimenot{0}{0}=\boldmatvec0$,
				$\transitionmat\timenot{1} = \boldmatvec0$

				\STATE                    
				\textbf{for}~{$t=1,\ldots,\timenum$}{  }\textbf{do}\\
				\STATE\label{step:prediction}\quad                          
				$\signalestvec\timegiventimenot{\timeind}{\timeind-1}=\transitionmat\timenot{\timeind}
				\signalestvec\timegiventimenot{\timeind-1}{\timeind-1}
				$\\
				\STATE\label{step:predictionerror}\quad                         
				$
				\errormat\timegiventimenot{t}{t-1}=\transitionmat\timenot{\timeind}\errormat\timegiventimenot{t-1}{t-1}\transitionmat\transpose\timenot{\timeind}+\plantnoisekernelmat\timenot{t}$\\
				\STATE\label{step:gain}\quad                          $
				\kalmangainmat\timenot{t}
				=\errormat\timegiventimenot{t}{t-1}\samplemat\transpose\timenot{t}
				(\observationnoisevar\timescalarnot{\timeind}\identitymat 
				+\samplemat\timenot{t}\errormat\timegiventimenot{t}{t-1}\samplemat\transpose\timenot{t})^{-1}$\\
				\STATE\label{step:correction}\quad
				$  
				\signalestvec\timenot{t|t}=\signalestvec\timenot{t|t-1}+\kalmangainmat\timenot{t}(\observationvec\timenot{t}-\samplemat\timenot{t}\signalestvec\timenot{t|t-1})
				$\\
				\STATE\label{step:correctionerror}\quad
				$  
				\errormat\timegiventimenot{t}{t}=(\identitymat-\kalmangainmat\timenot{t}\samplemat\timenot{t})\errormat\timegiventimenot{t}{t-1}$
			\end{algorithmic}
			
			\textbf{Output:} 
			$\signalestvec\timegiventimenot{\timeind}{\timeind}$, $\timeind =
			1,\ldots,\timenum$;
			$\errormat\timenot{\timeind}$, $\timeind =
			1,\ldots,\timenum$.
			
		\end{minipage}
	\end{algorithm}

\subsection{Multi-kernel kriged Kalman filters}
\label{sec:multkr}  
The following section applies the KRR framework presented 
in \S\ref{sec:tirecon} to online data-adaptive estimators of 
$\truesignal\timenot{\timeind}$. Specifically, a spatio-temporal model is 
presented that 
judiciously captures the dynamics over space and time. Based on this  
model
the KRR criterion over time and space is formulated,   and an online 
algorithm is derived with 
affordable computational complexity, when the kernels are 
pre-selected. To 
bypass the need for selecting an appropriate kernel, this section discusses a 
data-adaptive multi-kernel learning extension of the KRR estimator that learns 
the optimal kernel 
``on-the-fly."

\subsubsection{Spatio-temporal models}
\label{sec:spatiotemp}

Consider modeling the 
dynamics 
of 
$\truesignal\timenot{\timeind}$ separately over time and space as
		$\signalfun(v_\vertexind,\timeind)= 
		\truesignalspatiocompfun(v_\vertexind,\timeind) 
		+\truesignalspatiotempcompfun(v_\vertexind,\timeind),$
		or in vector form
		\begin{align}
			\truesignal\timenot{\timeind} 
			=\truesignalspatiocomp\timenot{\timeind}+\truesignalspatiotempcomp 
			\timenot{\timeind} 
			\label{eq:decompkr}
		\end{align}
		where $\truesignalspatiocomp\timenot{\timeind}\define[ 
		\truesignalspatiocompfun(v_1,\timeind),\ldots, 
		\truesignalspatiocompfun(v_\vertexnum,\timeind)]\transpose$ and 
		$\truesignalspatiotempcomp\timenot{\timeind}\define[ 
		\truesignalspatiotempcompfun(v_1,\timeind),\ldots, 
		\truesignalspatiotempcompfun(v_\vertexnum,\timeind)]\transpose$. The 
		first 
		term 
		$\{\truesignalspatiocomp\timenot{\timeind}\}_\timeind$ captures 
		only spatial dependencies, and can be thought of as 
		exogenous input to the graph that does not affect the 
		evolution of the function in time. 
		
		The second term
		$\{\truesignalspatiotempcomp\timenot{\timeind}\}_\timeind$ accounts for 
		spatio-temporal
		dynamics. 
A popular approach ~\cite[Ch. 
			3]{anderson1958} models 
			$\truesignalspatiotempcomp\timenot{\timeind}$ with the
			state equation
			\begin{align}
				\truesignalspatiotempcomp\timenot{\timeind}=	
				\adjtransgraphmat\timetimenot{\timeind}{\timeind-1}\truesignalspatiotempcomp\timenot{\timeind-1}
				+ \plantnoisevec\timenot{\timeind}, \quad\timeind=1,2,\ldots
				\label{eq:transmod}
			\end{align}
			where $\adjtransgraphmat\timetimenot{\timeind}{\timeind-1}$ is a 
			generic transition matrix that 
			can be chosen e.g. as the  
			$\vertexnum\times\vertexnum$ adjacency of a possibly directed 
			``transition graph,'' with 
			$\truesignalspatiotempcomp\timenot{0}=\boldmatvec 0$, and 
			$\plantnoisevec\timenot{\timeind}\in\rfield^{\vertexnum}$ capturing 
			the state error.
			The state transition matrix 
			$\adjtransgraphmat\timetimenot{\timeind}{\timeind-1}$ can be 
			selected in accordance with 
			the prior 
			information available. Simplicity in 
			estimation motivates the 
			random walk 
			model~\cite{rajawat2014cartography}, where 
			$\adjtransgraphmat\timetimenot{\timeind} 
			{\timeind-1}=\transweight\boldmatvec
			I_N$ 
			with $\transweight>0$. On the other hand, adherence to the graph, 
			prompts the selection 
			$\adjtransgraphmat\timetimenot{\timeind}{\timeind-1}= 
			\transweight 
			\adjacencymat$, in which 
			case~\eqref{eq:transmod} 
			amounts to a diffusion process on the time-invariant graph 
			$\graph$. The 
			recursion 
			in~\eqref{eq:transmod} is a vector 
			autoregressive model (VARM) of order one, 
			and offers flexibility in 
			tracking multiple 
			forms of 
			temporal 
			dynamics~\cite[Ch. 
			3]{anderson1958}. The model in~\eqref{eq:transmod} captures the 
			dependence between
			$\truesignalspatiotempcompfun(v_\vertexind,\timeind)$ and
			its time lagged versions 
			$\{\truesignalspatiotempcompfun(v_\vertexind,\timeind-1)\} 
			_{\vertexind=1}^\vertexnum$. 
			
			Next, a model with increased 
			flexibility is presented to  account for 
			instantaneous spatial dependencies as well 	
			\begin{align}
				%
				\truesignalspatiotempcomp\timenot{\timeind}= 
				\adjtransgraphmat\timetimenot{\timeind}{\timeind} 
				\truesignalspatiotempcomp\timenot{\timeind}+
				\adjtransgraphmat\timetimenot{\timeind}{\timeind-1} 
				\truesignalspatiotempcomp\timenot{\timeind-1}
				+ \plantnoisevec\timenot{\timeind},
				\quad\timeind=1,2,\ldots
				\label{eq:transmod3}
			\end{align}
			where $	\adjtransgraphmat\timetimenot{\timeind}{\timeind}$ encodes 
			the instantaneous 
			relation between 	
			$\truesignalspatiotempcompfun(v_\vertexind,\timeind)$ and
			$\{\truesignalspatiotempcompfun(v_\vertexindp,\timeind)\}_{\vertexindp\ne\vertexind}$. 
			The recursion in \eqref{eq:transmod3} 
			amounts to a structural vector autoregressive 
			model (SVARM)~\cite{shen2016nonlinear}. 
			\cmt{equivalence}Interestingly, \eqref{eq:transmod3} 
			can  be rewritten as 
			\begin{equation}
			 	\truesignalspatiotempcomp\timenot{\timeind}= 
			(\identitymat_\vertexnum-\adjtransgraphmat\timetimenot{\timeind}{\timeind})\inv
			\adjtransgraphmat\timetimenot{\timeind}{\timeind-1} 
			\truesignalspatiotempcomp\timenot{\timeind-1} 
			+ 	
			(\identitymat_\vertexnum-\adjtransgraphmat 
			\timetimenot{\timeind}{\timeind})\inv
			\plantnoisevec\timenot{\timeind}
			\end{equation}
			where $\identitymat_\vertexnum-\adjtransgraphmat\timetimenot 
			{\timeind}{\timeind}$ is assumed invertible. After defining 
			$\plantnoisetildvec\timenot{\timeind} 
			\define(\identitymat_\vertexnum-\adjtransgraphmat
			\timetimenot{\timeind}{\timeind})\inv
			\plantnoisevec\timenot{\timeind}$ and 
			$\adjacencytildmat\timetimenot{\timeind} 
			{\timeind-1}\define(\identitymat_\vertexnum-
			\adjtransgraphmat\timetimenot{\timeind}{\timeind})\inv
			\adjtransgraphmat\timetimenot{\timeind}{\timeind-1} $,  
			\eqref{eq:transmod3}  boils down to \begin{equation}
			    \truesignalspatiotempcomp\timenot{\timeind}= 
			\adjacencytildmat\timetimenot{\timeind}{\timeind-1}
			\truesignalspatiotempcomp\timenot{\timeind-1} 
			+ 	
			\plantnoisetildvec\timenot{\timeind} 
			\end{equation}
which is equivalent to~\eqref{eq:transmod}. This section will 
			focus on deriving estimators based on~\eqref{eq:transmod}, but can also 
			accommodate~\eqref{eq:transmod3} using the 
			aforementioned 
			reformulation.

	Modeling $\truesignal\timenot{\timeind}$ as the 
	superposition of a term $\truesignalspatiotempcomp \timenot{\timeind}$ 
	capturing
	the slow dynamics over time with a state space equation, and a term 
	$\truesignalspatiocomp \timenot{\timeind}$ accounting for fast dynamics is 
	motivated by the application at 
	hand~\cite{wikle1999dimension,rajawat2014cartography,kim2011kalmansensing}. In
	the 
	kriging 
	terminology~\cite{wikle1999dimension}, 
	$\truesignalspatiocomp\timenot{\timeind}$  is said to model small-scale 
	\emph{spatial fluctuations}, whereas 
	$\truesignalspatiotempcomp\timenot{\timeind}$ captures
	the so-called \emph{trend}. 
	The decomposition \eqref{eq:decompkr}
	is often dictated by the sampling interval: while 
	$\truesignalspatiotempcomp\timenot{\timeind}$ captures 
	slow dynamics relative to the sampling 	interval, fast variations are 
	modeled with	
	$\truesignalspatiocomp\timenot{\timeind}$. Such a modeling approach is motivated in the 
	prediction of network 
		delays~\cite{rajawat2014cartography}, where 
		$\truesignalspatiotempcomp\timenot{\timeind}$ 
		represents the 	queuing delay while 
		$\truesignalspatiocomp\timenot{\timeind}$ the 
		propagation, 
		transmission, 
		and processing delays. Likewise, when predicting prices 
		across different stocks,	
		$\truesignalspatiotempcomp\timenot{\timeind}$ captures the daily 
		evolution of the stock 
		market,  
		which is correlated across stocks and time 
		samples, while $\truesignalspatiocomp\timenot{\timeind}$ describes 
		unexpected changes, 
		such as the daily drop of the stock market due to political 
		statements, which 
		are	assumed uncorrelated over time.

\subsubsection{Kernel kriged Kalman filter}\label{sec:kkrkf}

	The 
	spatio-temporal model in~\eqref{eq:decompkr}, \eqref{eq:transmod} 
	can represent multiple forms of spatio-temporal dynamics by judicious 
	selection of the associated parameters. The 
	\emph{batch} KRR estimator over time 
	yields 
	\begin{align}
		%
		\underset{\{
			\truesignalspatiotempcomp\timenot{\tau},\plantnoisevec\timenot{\tau},
			\truesignalspatiocomp\timenot{\tau},\truesignal\timenot{\tau}
			\}_{\tau=1}^\timeind}{\argmin}&
		~\sum_{\tau=1}^{\timeind}\tfrac{1}{\samplenum\timescalarnot 
			{\tau}}\|\observationvec\timenot{\tau}
		-
		\samplemat\timenot{\tau}\truesignal\timenot{\tau}
		\|^2 \nonumber
		+\regparone\sum_{\tau=1}^{\timeind} 
		\|\plantnoisevec\timenot{\tau}\|^2_ 
		{\fullkernelstatenoisemat\timenot{\tau}}
		+\regpartwo\sum_{\tau=1}^{\timeind}\ 
		\|\truesignalspatiocomp\timenot{\tau}\|^2_ 
		{\fullkernelspatiomat\timenot{\tau}}\nonumber\\
		\subjectto~~~
		&\plantnoisevec\timenot{\tau}=\truesignalspatiotempcomp\timenot{\tau}-
		\adjtransgraphmat 
		\timetimenot{\tau}{\tau-1}\truesignalspatiotempcomp 
		\timenot{\tau-1},~ 
		\truesignal\timenot{\tau}=\truesignalspatiocomp\timenot{\tau} + 
		\truesignalspatiotempcomp\timenot{\tau},~~\tau=1,\ldots,\timeind. 
		\label{eq:rkhsobjinit}		
	\end{align}
	The first term in~\eqref{eq:rkhsobjinit} penalizes the fitting 
	error in accordance 
	with~\eqref{eq:observationsvec}. The scalars 
	$\regparone,\regpartwo\ge0$ are regularization parameters 
	controlling the effect  of the kernel regularizers, while prior 
	information about 
	$\{\truesignalspatiocomp\timenot{\tau}, 
	\plantnoisevec\timenot{\tau}\}_ 
	{\tau=1}^\timeind$ may guide the selection of the appropriate 
	kernel matrices. The constraints 
	in~\eqref{eq:rkhsobjinit} imply adherence to~\eqref{eq:transmod} 
	and~\eqref{eq:decompkr}.	
	Since the $\truesignalspatiocomp\timenot{\tau}, 
	\plantnoisevec\timenot{\tau}$ are defined over the 
	time-evolving 
	$\graph\timescalarnot{\tau}$, a potential approach is to select 
	Laplacian kernels as 
	$\fullkernelspatiomat\timenot{\tau},
	\fullkernelstatenoisemat\timenot{\tau}$, see 
	\S\ref{sec:kernelsong}. 
	Next, we rewrite \eqref{eq:rkhsobjinit} in a form amenable to online solvers, namely 
	\begin{eqnarray}
		%
		\underset{\{
			\truesignalspatiotempcomp\timenot{\tau},
			\truesignalspatiocomp\timenot{\tau}
			\}_{\tau=1}^\timeind}{\argmin} & &
		~\sum_{\tau=1}^{\timeind}\tfrac{1}{\samplenum\timescalarnot 
			{\tau}}\|\observationvec\timenot{\tau}
		-
		\samplemat\timenot{\tau}\truesignalspatiotempcomp\timenot{\tau}
		-\samplemat\timenot{\tau}\truesignalspatiocomp\timenot{\tau}	  
		\|^2 +
		\nonumber \\
		& &+\regparone\sum_{\tau=1}^{\timeind}\|\truesignalspatiotempcomp\timenot{\tau}-
		~\adjtransgraphmat\timetimenot{\tau}{\tau-1} 
		\truesignalspatiotempcomp\timenot
		{\tau-1}\|^2_{\fullkernelstatenoisemat\timenot{\tau}}+
 		\nonumber\\ 		
 		& & +
		\regpartwo\sum_{\tau=1}^ 
		{\timeind}\|\truesignalspatiocomp\timenot{\tau}\|^2_ 
		{\fullkernelspatiomat\timenot{\tau}} 
		\label{eq:rkhsobj}.	
	\end{eqnarray}
	
	In a batch form the  optimization in~\eqref{eq:rkhsobj} yields  
	$		 
	\{\estsignalspatiocomp\timenot{\tau|\timeind}$, and $
	\estsignalspatiotempcomp\timenot{\tau|\timeind}\}_{\tau=1}^\timeind
	$ per slot $t$ with complexity that grows with $\timeind$. Fortunately, the 
	\emph{filtered} solutions 
	$\{\estsignalspatiocomp\timenot{\tau|\tau}, 
	\estsignalspatiotempcomp\timenot{\tau|\tau} 
	\}_{\tau=1}^\timeind$ of~\eqref{eq:rkhsobj},  are attained by the 
	kernel kriged Kalman 
	filter 
	(KeKriKF) in an \emph{online} 
	fashion. For 
	the proof the reader is referred to~\cite{ioannidis2017kriged}.
	One iteration of the proposed KeKriKF is summarized  as 
	Algorithm~\ref{algo:krkalmanfilter}. This online estimator -- with 
		computational complexity $\mathcal{O}(\vertexnum^3)$ per $\timeind$ --  
		tracks the
		temporal variations of the signal of interest 
		through~\eqref{eq:transmod}, and promotes 
		desired properties 
		such as smoothness over the graph, using 
		$\fullkernelspatiomat\timenot{\timeind}$. 
		and 
		$\fullkernelstatenoisemat\timenot{\timeind}$. Different from 
			existing KriKF approaches over 
			graphs~\cite{rajawat2014cartography}, the KeKriKF takes 
			into account 
			the 
			underlying  graph structure in estimating
			$\truesignalspatiocomp\timenot{\timeind}$ as well as 
			$\truesignalspatiotempcomp\timenot{\timeind}$. Furthermore, by using 
			$\laplacianmat\timenot{\timeind}$
			in~\eqref{eq:laplacian_kernel}, it can also accommodate dynamic 
			graph topologies. Finally, it should be noted that 
			KeKriKF encompasses 
			as a special case 
			the KriKF, which relies on knowing the statistical properties of the function~\cite{rajawat2014cartography,wikle1999dimension,kim2011kalmansensing, 
				mardia1998kriged}. 
	
	Lack of prior information prompts the development of data-driven approaches that 
	efficiently learn the appropriate kernel matrix. In the next section, we discuss an online MKL approach for achieving this goal.
	\begin{algorithm}[t]                
		\caption{Kernel Kriged Kalman filter (KeKriKF)}
		\label{algo:krkalmanfilter}
		\indent\textbf{Input:} 
		$\fullkernelstatenoisemat\timenot{\timeind}; 
		\fullkernelspatiomat\timenot{\timeind}
		\in\pdset^{\vertexnum}$;
		$\adjtransgraphmat\timetimenot{\timeind}{\timeind-1}\in 
		\rfield^{\vertexnum\times\vertexnum}$; 
		$\observationvec\timenot{\timeind}\in 
		\rfield^{\samplenum\timescalarnot{\timeind}}$;
		$\samplemat\timenot{\timeind}\in\{0,1\}^ 
		{\samplenum\timescalarnot{\timeind}\times\vertexnum}$.
		\\  
		\begin{algorithmic}[1]  
			\STATE$
			\genkernelmat\timenot{\timeind}=\frac{1}{\regpartwo} 
			\samplemat\timenot{\timeind}\fullkernelspatiomat\timenot{\timeind} 
			\samplemat\transpose\timenot{\timeind}+ 
			\samplenum\timescalarnot{\timeind}\identitymat_
			{\samplenum\timescalarnot{\timeind}}$\\
			\STATE$\estsignalspatiotempcomp 
			\timegiventimenot{\timeind}{\timeind-1}=\adjtransgraphmat\timetimenot{\timeind}{\timeind-1}
			\estsignalspatiotempcomp\timegiventimenot{\timeind-1}{\timeind-1}
			$\\
			\STATE$
			\errormat\timegiventimenot{t}{t-1}= 
			\adjtransgraphmat\timetimenot{\timeind}{\timeind-1}
			\errormat
			\timegiventimenot{t-1}{t-1}\adjtransgraphmat\transpose\timetimenot{\timeind}{\timeind-1}+
			\frac{1}{\regparone}\fullkernelstatenoisemat\timenot{t}$\\
			\STATE$
			\kalmangainmat\timenot{t}
			=\errormat\timegiventimenot{t}{t-1} 
			\samplemat\transpose\timenot{t}(
			\genkernelmat\timenot{\timeind}
			+\samplemat\timenot{t}\errormat\timegiventimenot{t}{t-1}
			\samplemat\transpose\timenot{t})^{-1}$\\
			\STATE$  
			\errormat\timegiventimenot{t}{t}=(\identitymat-\kalmangainmat 
			\timenot{t}\samplemat\timenot{t})\errormat\timegiventimenot{t}{t-1}$\\
			\STATE$  
			\estsignalspatiotempcomp\timenot{t|t}= 
			\estsignalspatiotempcomp\timenot{t|t-1}+		 
			\kalmangainmat\timenot{t}(\observationvec\timenot{t}- 
			\samplemat\timenot{t}\estsignalspatiotempcomp\timenot{t|t-1})
			$
			\\
			\STATE$
			\estsignalspatiocomp\timegiventimenot{\timeind}{\timeind}= 
			\fullkernelspatiomat\timenot{\timeind} 
			\samplemat\transpose\timenot{\timeind} 
			{\genkernelmat\timenot{\timeind}}\inv
			(\observationvec\timenot{\timeind} 
			-\samplemat\timenot{\timeind}\estsignalspatiotempcomp\timenot{\timeind|\timeind})
			$\\			
		\end{algorithmic}
		\indent\textbf{Output:} 
		$\estsignalspatiotempcomp\timegiventimenot{\timeind}{\timeind}$;
		$\estsignalspatiocomp\timegiventimenot{\timeind}{\timeind}$;
		$\errormat\timegiventimenot{\timeind}{\timeind}$.
	\end{algorithm} 
\subsubsection{Online multi-kernel Learning}
\label{sec:omk}

	To cope with lack of prior information about the pertinent kernel, the 
	following dictionaries of kernels will be considered
	$\fullkernelspatiomatdict\define 
	\{\fullkernelspatiomat\kernelindnot{\rkhsind}
	\in\pdset^{\vertexnum}\}_{\rkhsind=1}^\rkhsspationum$ and 
	$\fullkernelstatenoisematdict\define\{\fullkernelstatenoisemat\kernelindnot{\rkhsind}
	\in\pdset^{\vertexnum}\}_{\rkhsind=1}^\rkhsstatenoisenum$.
	For the following assume that 
	$\fullkernelspatiomat\timenot{\tau}=\fullkernelspatiomat$,
	$\fullkernelstatenoisemat\timenot{\tau}=\fullkernelstatenoisemat$ and 
	$\samplemat\timenot{\tau}=\samplemat ,~\forall\tau$. Moreover, we postulate  
	that the kernel matrices are of the form 
	$\fullkernelspatiomat=\fullkernelspatiomat 
	\kernelcoefnot{\kernelcoefspatiovec}=
	\sum_{\rkhsind=1}^{\rkhsspationum}\kernelcoefspatio\kernelindnot{\rkhsind} 
	\fullkernelspatiomat\kernelindnot{\rkhsind}$ and 
	$\fullkernelstatenoisemat=\fullkernelstatenoisemat 
	\kernelcoefnot{\kernelcoefstatenoisevec}= 
	\sum_{\rkhsind=1}^{\rkhsstatenoisenum} 
	\kernelcoefstatenoise\kernelindnot{\rkhsind} 
	\fullkernelstatenoisemat\kernelindnot{\rkhsind}$, where 
	$\kernelcoefstatenoise\kernelindnot{\rkhsind} , 
	\kernelcoefspatio\kernelindnot{\rkhsind}  \ge   0,~\forall \rkhsind $.  
	
	Next, in 
	accordance with \S\ref{sec:select_kernels} the coefficients 
	$\kernelcoefspatiovec=[\kernelcoefspatio\kernelindnot{1},\ldots, 
	\kernelcoefspatio\kernelindnot{\rkhsnum}]\transpose$ and 
	$\kernelcoefstatenoisevec=[\kernelcoefstatenoise\kernelindnot{1},\ldots, 
	\kernelcoefstatenoise\kernelindnot{\rkhsnum}]\transpose$ can be found by 
	jointly
	minimizing~\eqref{eq:rkhsobj} with respect to 
	$\{
	\truesignalspatiotempcomp\timenot{\tau},
	\truesignalspatiocomp\timenot{\tau}
	\}_{\tau=1}^\timeind, \kernelcoefspatiovec$ and $
	\kernelcoefstatenoisevec$ that yields 
	\begin{align}
		%
		\underset{~~~\{
			\truesignalspatiotempcomp\timenot{\tau},
			\truesignalspatiocomp\timenot{\tau}
			\}_{\tau=1}^\timeind, \atop\kernelcoefspatiovec\ge \boldmatvec  0 , 
			\kernelcoefstatenoisevec\ge  \boldmatvec 0 }{\argmin}&
		~\sum_{\tau=1}^{\timeind}\tfrac{1}{\samplenum}\|\observationvec\timenot{\tau}
		-
		\samplemat\truesignalspatiotempcomp\timenot{\tau}
		-\samplemat\truesignalspatiocomp\timenot{\tau}	  
		\|^2 
		+\regparone\sum_{\tau=1}^{\timeind}\|\truesignalspatiotempcomp\timenot{\tau}-
		~\adjtransgraphmat\timetimenot{\tau}{\tau-1} 
		\truesignalspatiotempcomp\timenot
		{\tau-1}\|^2_{\fullkernelstatenoisemat\kernelcoefnot{\kernelcoefstatenoisevec}}
		\nonumber\\
		+&\regpartwo\sum_{\tau=1}^{\timeind} 
		\|\truesignalspatiocomp\timenot{\tau}\|^2_
		{\fullkernelspatiomat\kernelcoefnot{\kernelcoefspatiovec}}+ 
		\timeind\kernelcoefspatioreg\|\kernelcoefspatiovec\|_2^2
		+\timeind\kernelcoefstatenoisereg\|\kernelcoefstatenoisevec\|_2^2
		\label{eq:multrkhsobj}		
	\end{align}
	where $\kernelcoefspatioreg, \kernelcoefstatenoisereg\ge0$ are 
	regularization 
	parameters, that effect a ball constraint on $\kernelcoefspatiovec$ and 
	$\kernelcoefstatenoisevec$, weighted
	by $\timeind$ to account for the first three terms that are growing with 
	$\timeind$. Observe that the optimization problem in~\eqref{eq:multrkhsobj} 
	gives time varying 
	estimates $\kernelcoefspatiovec\timenot{\timeind}$ and  
	$\kernelcoefstatenoisevec\timenot{\timeind}$ 
	allowing to track the optimal $\fullkernelspatiomat$ and 
	$\fullkernelstatenoisemat$ that 
	change over time respectively.

	The optimization problem 
		in~\eqref{eq:multrkhsobj} is not jointly 
		convex in $\{
		\truesignalspatiotempcomp\timenot{\tau},
		\truesignalspatiocomp\timenot{\tau}
		\}_{\tau=1}^\timeind, \kernelcoefspatiovec, 
		\kernelcoefstatenoisevec$, but it is separately convex in these 
		variables. To 
		solve~\eqref{eq:multrkhsobj} alternating minimization
		 strategies will be employed, that suggest optimizing with respect 
		to one 
		variable, while keeping the other variables 
		fixed~\cite{csisz1984information}. If 
		$\kernelcoefspatiovec, 
		\kernelcoefstatenoisevec$ are considered fixed, \eqref{eq:multrkhsobj} 
		reduces to \eqref{eq:rkhsobj}, 
		which can be solved by
		Algorithm~\ref{algo:krkalmanfilter} for the estimates 
		$\estsignalspatiotempcomp\timegiventimenot{\timeind}{\timeind}, 
		\estsignalspatiocomp\timegiventimenot{\timeind}{\timeind}$ at each 
		$\timeind$. For $\{
		\truesignalspatiotempcomp\timenot{\tau},
		\truesignalspatiocomp\timenot{\tau}
		\}_{\tau=1}^\timeind$ fixed and replaced by $\{
		\estsignalspatiotempcomp\timegiventimenot{\tau}{\tau}, 
		\estsignalspatiocomp\timegiventimenot{\tau}{\tau}
		\}_{\tau=1}^\timeind$ in~\eqref{eq:rkhsobj} the time-varying estimates 
		of
		$\kernelcoefspatiovec,\kernelcoefstatenoisevec$ are found by
		\begin{subequations}
			\begin{align}
				%
				\kernelcoefstatenoisevecest\timenot{\timeind}&=\underset{\kernelcoefstatenoisevec\ge
					\boldmatvec 
					0}{\argmin}
				\tfrac{1}{\timeind}\sum_{\tau=1}^{\timeind} 
				\|\estsignalspatiotempcomp\timegiventimenot{\tau}{\tau}-
				~\adjtransgraphmat\timetimenot{\tau}{\tau-1} 
				\estsignalspatiotempcomp\timegiventimenot{\tau-1}{\tau-1} 
				\|^2_{\fullkernelstatenoisemat\kernelcoefnot{\kernelcoefstatenoisevec}}
				+\tfrac{\kernelcoefstatenoisereg}{\regparone}\|\kernelcoefstatenoisevec\|_2^2
				\label{eq:stateermultker}\\
				\kernelcoefspatiovecest\timenot{\timeind}&=\underset{\kernelcoefspatiovec\ge
					\boldmatvec 0}{\argmin} 
				\tfrac{1}{\timeind}\sum_{\tau=1}^{\timeind} 
				\|\estsignalspatiocomp\timegiventimenot{\tau}{\tau}\|^2_
				{\fullkernelspatiomat\kernelcoefnot{\kernelcoefspatiovec}}+ 
				+\tfrac{\kernelcoefspatioreg}{\regpartwo} 
				\|\kernelcoefspatiovec\|_2^2.
				\label{eq:spatiomultker}
			\end{align}
		\end{subequations}

		 The optimization
		problems~\eqref{eq:stateermultker} and~\eqref{eq:spatiomultker}
		 are strongly convex and iterative algorithms 
		are available based on projected gradient 
		descent (PGD)~\cite{zhang2016multikernelssp}, or the Frank-Wolfe algorithm~\cite{tsp2016zwrg}. When the 
		kernel 
		matrices belong to the 
		Laplacian family~\eqref{eq:laplacian_kernel}, efficient algorithms that 
		exploit the common eigenspace of the kernels in the dictionary have 
		been developed in~\cite{ioannidis2017kriged}. The proposed method 
		reduces the per step computational 
		complexity of PGD from a prohibitive 
		$\mathcal{O}(\vertexnum^{3}\rkhsnum)$ for general kernels to a more  
		affordable 
		$\mathcal{O}(\vertexnum\rkhsnum)$ for Laplacian kernels.
		The proposed algorithm, termed multi-kernel KriKF (MKriKF) alternates between computing  
 $\estsignalspatiotempcomp\timegiventimenot{\timeind}{\timeind}$ 
and $\estsignalspatiocomp\timegiventimenot{\timeind}{\timeind}$ utilizing the KKriKF and estimating $ 
\kernelcoefspatiovecest\timenot{\timeind}$ and $ 
\kernelcoefstatenoisevecest\timenot{\timeind}$ from
solving~\eqref{eq:spatiomultker} and~\eqref{eq:stateermultker}.

\newcommand{\dlsrstepsize}{\hc{\mu_\text{DLSR}}}
\newcommand{\lmsstepsize}{\hc{\mu_\text{LMS}}}
\newcommand{\dlsrbeta}{\hc{\beta_\text{DLSR}}}
\newcommand{\changev}[1]{\textcolor{green}{#1}}

\subsection{Numerical tests}
\cmt{Sec. overview}This section compares the performance of the
methods we discussed in \S\ref{sec:reconstruction} and  \S\ref{sec:multkr} with state-of-the-art 
alternatives, 
and illustrates
some of the trade-offs inherent to time-varying function
reconstruction through real-data experiments. The source code for the
simulations is available at the authors' websites.

 Unless otherwise stated, the compared estimators include
  distributed least squares reconstruction
	(DLSR)~\cite{wang2015distributed} with step size $\dlsrstepsize$ and
	parameter $\dlsrbeta$;  the least mean-squares (LMS) algorithm
	in~\cite{lorenzo2016lms} with step size $\lmsstepsize$; 
	the bandlimited instantaneous estimator (BL-IE), which results after 
	applying~\cite{narang2013localized,tsitsvero2016uncertainty,anis2016proxies}
	separately per $\timeind$;  and the KRR instantaneous estimator
	(KRR-IE)  in \eqref{eq:timeagnostic} with a diffusion
	kernel with parameter $\sigma$. 
DLSR, LMS, and BL-IE also use a bandwidth parameter~$\bandwidth$.

\noindent\textbf{Reconstruction via extended graphs.} {      
	For our first experiment we use a dataset  obtained from  an epilepsy 
	study~\cite{kramer2008seizure}, which is used to showcase an example analysis of   
	electrocorticography (ECoG) data (analysis of ECoG data is a standard tool in diagnosing epilepsy). 
}
Our next experiment utilizes the 
	ECoG time series  in~\cite{kramer2008seizure} from $\vertexnum=76$
	electrodes implanted in a patient's brain before and after
	the onset of a seizure.
 A symmetric time-invariant
		adjacency matrix $\spaceadjacencymat$ was obtained
		using the method in~\cite{shen2016nonlinear} with ECoG
		data before the onset of the seizure.
		Function
		$\signalfun\timevertexnot{\timeind}{\vertexind}$
		comprises the electrical signal at the $\vertexind$-th
		electrode and $\timeind$-th sampling instant after the
		onset of the seizure, for a period of $\timenum=250$
		samples. The values of
		$\signalfun\timevertexnot{\timeind}{\vertexind}$ were
		normalized by subtracting the temporal mean of each
		time series  before the onset of the
		seizure.
	The goal of the experiment is to illustrate the  
	reconstruction performance of  KKF in
	capturing the complex spatio-temporal dynamics of
	brain signals.
	\setlength{\mywidth}{.40\textwidth}
    \setlength{\myheight}{.18\textwidth}
		\begin{figure}
		\centering
		\subfigure[NMSE for the ECoG data set
			($\sigma=1.2$, $\regpar=10^{-4}$, $\dlsrstepsize=1.2$, $\timeconnection =0.01$,
			$\dlsrbeta =0.5$, $\lmsstepsize =0.6$).
			]{
			\includegraphics[width=.47\linewidth]
			{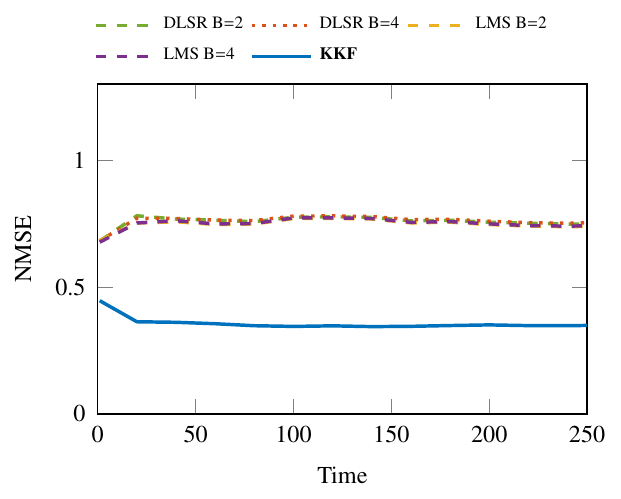}
			}
 			\hfill
		\subfigure[NMSE of temperature estimates. 
		($\regparone=1$, 
			$\regpartwo=1$, $\dlsrstepsize
			=1.6$, $\dlsrbeta=0.5$, $\lmsstepsize =0.6$, 
			$\transweight=10^{-3}$, 
			$\gausmeanstatenoise=10^{-5}$, $\gausstdstatenoise=10^{-6}$, 
			$\gausmeanspatio=2$, 
			$\gausstdspatio=0.5$, $\rkhsspationum=40$, 
			$\rkhsstatenoisenum=40$)]{\includegraphics[width=.47\linewidth]
				{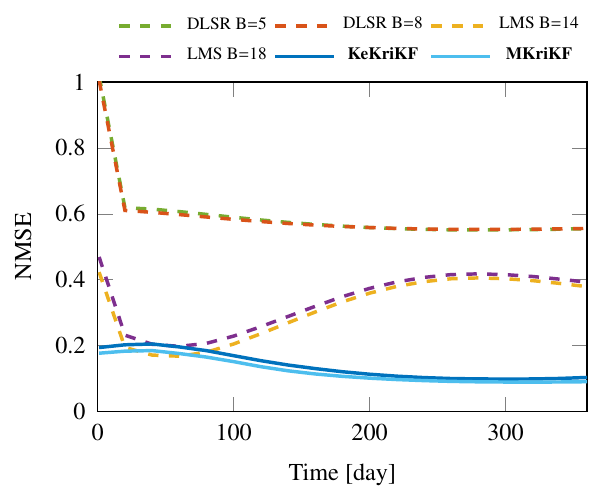}}
 			\hfill
%
		\caption{NMSE for real data simulations.} 
		\label{fig:brainsignalacrosstime}
	\end{figure}
	
	{ Fig.~\ref{fig:brainsignalacrosstime}(a) depicts the
		$\text{NMSE}(\timeind,\{\sampleset\timenot{\tau}\}_{\tau=1}^\timeind)$,
		averaged over all sets
		$\sampleset\timenot\timeind=\sampleset,~\forall\timeind,$ of size
		$\samplenum=53$.  For the KKF, a
		space-time kernel was created (see~\cite{romero2016spacetimekernel})
		with $\spacefullkernelmat\timenot{\timeind}$ a time-invariant
		covariance kernel 
		$\spacefullkernelmat\timenot{\timeind}=\hat{\boldmatvec
			\Sigma}$, where $\hat{\boldmatvec \Sigma}$ was set to
		the sample covariance matrix of the time series before the onset of
		the seizure, and with a time-invariant $\timeconnectionmat=
		\timeconnection\boldmatvec I$.
		The results clearly show the superior reconstruction performance of
		the KKF, which successfully exploits the statistics of
		the signal when available, among competing approaches,
		even with a small number of samples. This result
		suggests that the ECoG diagnosis technique could be
		efficiently conducted even with a smaller number of
		intracranial electrodes, which may have a posite impact on 
		the patient's experience.  }

\noindent\textbf{Reconstruction via KeKriKF.} 
	The second dataset is provided by the  National Climatic Data 
	Center~\cite{USATemp}, and 
	comprises  
	hourly temperature measuments at $\vertexnum=109$ measuring stations 
	across the continental 
	United States in 2010. 
	A time-invariant graph was 
		constructed  as in 
		\cite{romero2016spacetimekernel}, based on geographical distances. 
		The value
		$\signalfun\timevertexnot{\timeind}{\vertexind}$ represents the
		temperature recorded at the $\vertexind$-th  station and
		$\timeind$-th day.
		
		Fig.~\ref{fig:brainsignalacrosstime}(b) reports  the performance of 
	different reconstruction 
	algorithms in terms of NMSE, for 
	$\samplenum=40$. The KeKriKF Algorithm~\ref{algo:krkalmanfilter} adopts a 
	diffusion kernel for 
	$\fullkernelspatiomat$ with 
	$\sigma=1.8$,  and for 
	$\fullkernelstatenoisemat=\plantnoiseweight\identitymat_\vertexnum$ 
	with $\plantnoiseweight=10^{-5}$. The multi-kernel kriged Kalman filter 
	(MKriKF) is configured with: 		$\fullkernelspatiomatdict$ that 
		contains $\rkhsspationum$ 
		diffusion kernels with parameters 
		$\{\sigma\kernelindnot{\rkhsind}\}_{\rkhsind=1}^\rkhsspationum$ drawn 
		from a Gaussian 
		distribution with mean $\gausmeanspatio$ and variance $\gausstdspatio$;
		$\fullkernelstatenoisematdict$ that contains $\rkhsstatenoisenum$ 
		$\plantnoiseweight\identitymat_\vertexnum$ with parameters 
		$\{\plantnoiseweight\kernelindnot{\rkhsind}\}_{\rkhsind=1}^\rkhsstatenoisenum$
		drawn 
		from a 
		Gaussian 
		distribution with mean $\gausmeanstatenoise$ and variance 
		$\gausstdstatenoise$. 
	The specific kernel selection for KeKriKF leads to the smallest NMSE error and 
were selected using cross validation. Observe that MKriKF captures the 
	spatio-temporal dynamics, 
	successfully explores the pool of available kernels, and
	achieves superior performance.

		\setlength{\mywidth}{.85\textwidth}
    \setlength{\myheight}{.18\textwidth}
		\begin{figure}
		\centering
		\subfigure[Tracking of GDP. ($\dlsrstepsize
					=1.6$, $\dlsrbeta=0.4$, $\lmsstepsize =1.6$,
					$\kernelcoefspatioreg=10^{5}$, 
					$\kernelcoefstatenoisereg=10^{5}$)]{
						\includegraphics
						{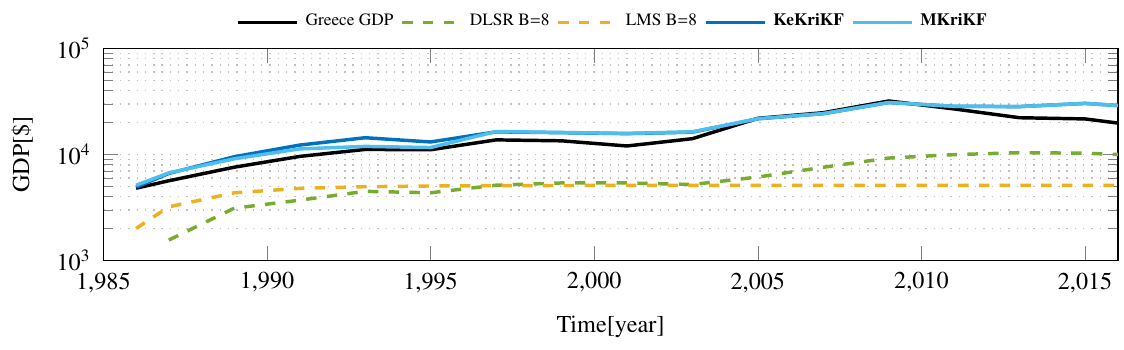}}
 			\hfill
		\caption{NMSE for GDP data.} 
		\label{fig:gdp}
	\end{figure}
	
		The  third dataset is provided by the World Bank Group~\cite{wordbank} and  comprises of  the 
		gross domestic product (GDP) per capita values for $\vertexnum=127$ countries  for the years 
		1960-2016. 
		A time-invariant graph was constructed using the correlation between the GDP values for the first 
		25 years 
		of different countries.  The graph function
			$\signalfun\timevertexnot
			{\timeind}{\vertexind}$ denotes the
			GDP value reported at the $\vertexind$-th  country and
			$\timeind$-th year for $\timeind=1985,\ldots,2016$.
			The graph fourier transform of the GDP values shows that the graph frequencies
				$\fouriersignalfun_k$, $4<k<120$ take
				small values and large values otherwise.
				Motivated by the aforementioned observation, the KKriKF is configured with a band-reject 
				kernel
				$\fullkernelspatiomat$ that 
				results after applying $\frequencyweightfun(\lambda_\vertexind)=\beta$ for 
				$k\leq\vertexind\leq \vertexnum-l$ and $\frequencyweightfun(\lambda_\vertexind)=1/\beta$ 
				otherwise 
				in~\eqref{eq:laplacian_kernel} with $k=3, l=6, \beta=15$ and for 
				$\fullkernelstatenoisemat=\plantnoiseweight\identitymat_\vertexnum$ 
				with $\plantnoiseweight=10^{-4}$. The MKriKF adopts a $\fullkernelspatiomatdict$ that 
				contains 
				band-reject kernels with $k\in[2,4],$ $ l\in[3,6],$ $ \beta=15$ that result to 
				$\rkhsspationum=12$ 
				kernels and a
				$\fullkernelstatenoisematdict$ that contains
				$\{\plantnoiseweight\kernelindnot{\rkhsind} 
				\identitymat_\vertexnum\}_{\rkhsind=1}^{40}$
				with $\plantnoiseweight\kernelindnot{\rkhsind}$ drawn 
				from a 
				Gaussian 
				distribution with mean $\gausmeanstatenoise=10^{-5}$ and variance 
				$\gausstdstatenoise=10^{-6}$.
			Next, the performance of different algorithms in tracking the GDP value is 
			evaluated after sampling $\samplenum=38$ 
			countries.
		
			Fig.~\ref{fig:gdp} illustrates the actual GDP as well as GDP estimates for Greece, that is not 
			contained in the sampled countries.  Clearly, MKriKF, that learns the pertinent kernels from the data,  
		achieves  roughly the same
		 performance of KKriKF, that is configured manually to obtain the smallest possible NMSE.

\section{Summary}
The task of reconstructing functions defined on graphs arises naturally in a plethora of applications. The kernel-based approach offers a clear, principled and intuitive way for tackling this problem. In this chapter, we gave a contemporary treatment of this framework focusing on both time-invariant and time-evolving domains. The methods presented herein offer the potential of providing an expressive way to tackle interesting real-world problems. Besides illustrating the effectiveness of the discussed approaches, our tests were also chosen to showcase interesting application areas as well as reasonable modeling approaches for the interested readers to build upon. 
For further details about the models discussed here and their theoretical 
properties, the reader is referred 
to~\cite{romero2017kernel,ioannidis2016semipar,ioannidis2017kriged,
	ioannidis2016kkrkf,ioannidis2016spacetime,romero2016multikernelssp} and the 
	references therein. 

\noindent
{\bf Acknowledgement}. 
The 
research was supported by NSF grants 1442686, 1500713, 
1508993, 1509040, and 1739397.

\bibliographystyle{IEEEbib}
\bibliography{my_bibliography}

\end{document}